\documentclass[11pt]{article}
\usepackage[preprint]{acl}
\usepackage{times}
\usepackage[no-math]{fontspec}
\usepackage{amsmath,amssymb,mathtools,booktabs,tabularx,multirow,graphicx}
\usepackage{flafter,placeins,fvextra}
\usepackage{needspace}
\usepackage{flushend}
\usepackage{ragged2e}
\hypersetup{pdftitle={How Many Humans Are 32 LLM Judges Worth?},pdfauthor={Chao Li; Yingying Yu; Yunfeng Li},pdfsubject={EN1.1CH0.4 arXiv preprint},unicode=true}
\DefineVerbatimEnvironment{Prompt}{Verbatim}{breaklines=true,breakanywhere=true,fontsize=\small,frame=single}


\title{How Many Humans Are 32 LLM Judges Worth?}
\author{\begin{tabular}{@{}c@{\hspace{0.7cm}}c@{\hspace{0.7cm}}c@{}}
  Chao Li\thanks{Corresponding author.} & Yingying Yu & Yunfeng Li \\[-1pt]
  \textnormal{Tsinghua University} &
  \textnormal{University College London} &
  \textnormal{\begin{tabular}[t]{@{}c@{}}Hebei University of\\Economics and Business\end{tabular}} \\[-1pt]
  \small\texttt{lichaocasia@gmail.com} &
  \small\texttt{yuyingying311@gmail.com} &
  \small\texttt{liyunfengxiaolong@gmail.com}
\end{tabular}}
\date{}
\begin{document}
\maketitle
\begin{abstract}
A panel's human-equivalent size is target-specific. Matching a fixed 32-judge panel to empirical human label distributions on three ChaosNLI tasks yields two distinct effective sizes: distributional-error matching gives $\nu_{\mathrm{MSE}}=2.304$, $3.750$, and $3.445$, whereas spectral matching gives $\nu_H=4.242$, $6.459$, and $6.499$, a gap of $1.72$--$1.89\times$; a binary-error diagnostic credits the same panels with only $1.971$--$2.227$ effective votes. Extrapolating the distributional-error curve at fixed squared mean residual, mean member variance, and normalized mean covariance gives asymptotes of $2.392$, $3.990$, and $3.655$, with 32 judges already reaching $94.0$--$96.3\%$. An exact spectral identity explains the gap: error depends on member energy and on the orientation of residual variation relative to averaging, information that the participation ratio (PR) discards. A realizable hard-label construction confirms that higher spectral diversity can coexist with worse distribution recovery even under equal member energies and nonnegative correlations, and the consensus direction retains $\gamma_{\mathrm{co}}=43.8\%$, $33.7\%$, and $35.9\%$ of centered residual variance. An external check on CC-1000, a 1,000-item Civil Comments subset with a different panel, gives $\nu_H=2.84$. For panel choice, we establish an existence result and one feasible path: exhaustive enumeration at $k\in\{5,7\}$ shows that panels beating the accuracy-top-$k$ baseline on both accuracy and $\nu_H$ always exist, and greedily swapping at most two members reaches $24.8$--$56.0\%$ higher $\nu_H$ at $0.10$--$1.10$ percentage points higher accuracy. Our dataset and code are available at \url{https://github.com/Chao1208/32judges-votes}.

\end{abstract}
\section{Introduction}\label{sec:intro}
Language models increasingly supply judgments for evaluation and data annotation. Their agreement with human judgments motivates using them in place of costly manual assessment \citep{zheng2023}, but every judge added to a panel also adds correlated responses. An effective-vote analysis found that nine judges could supply roughly two independent votes \citep{kohli2026b}. On three ChaosNLI tasks, we find that a 32-judge panel still yields only 1.971--2.227 binary-error effective votes, while its distributional-error effective size is 2.304--3.750. A panel's nominal size therefore says little about what its judgments contribute.

An effective count also needs an explicit target. Binary errors relative to a single gold label record whether a judge agrees with that label, but collapse the identities of all alternatives. If human label frequencies are $(0.6,0.3,0.1)$, choosing the second or third label incurs the same binary error while deviating from human judgments in different directions. Human disagreement in language inference has reproducible structure \citep{pavlick2019}, and label variation matters for evaluation \citep{plank2022}. Dense annotations in ChaosNLI \citep{nie2020} let us measure model residuals relative to that distribution.

This reference also distinguishes residual diversity from distribution recovery. A panel can span several residual directions while its average label frequencies remain far from the human distribution. Participation ratio (PR) summarizes spectral dimension \citep{mazzucato2016}; the squared error of an average also depends on member error energies and on the alignment of residual directions with averaging. Prior analyses study correlated model errors \citep{kim2025}, diversity versus majority-vote gains \citep{kim2026diversity}, and common errors retained by aggregation \citep{afrin2026}. We quantify these distinctions on the same answers and identify the energy and orientation information that spectral diversity discards.

The study covers a fixed pool of 32 models on three categorical language-inference tasks from ChaosNLI and one external check on CC-1000, a frozen 1,000-item Civil Comments toxicity subset with a different item domain, human-label design, and judge panel. Our target is each item's empirical human label distribution. We connect spectral size $\nu_H$, error-matched size $\nu_{\mathrm{MSE}}$, and binary-error effective votes $n_{\mathrm{eff}}$, explain their divergence, and find joint improvements in panel accuracy and effective size.

We make three contributions.
\begin{enumerate}
\RaggedRight
\item \textbf{Target-specific effective sizes expose sharply diminishing returns beyond 32 judges.} The same panels give $\nu_{\mathrm{MSE}}=2.304$, $3.750$, and $3.445$ against $\nu_H=4.242$, $6.459$, and $6.499$ (ratios of 1.841, 1.723, and 1.887), while the binary-error diagnostic stays near two votes. Extrapolating the distributional-error curve, holding the observed squared mean residual, mean member variance, and normalized mean covariance fixed, gives asymptotes of $2.392$, $3.990$, and $3.655$, with the 32-judge panels already reaching 94.0--96.3\%.
\item \textbf{An exact spectral identity explains the divergence between diversity and recovery.} The identity separates member energy and alignment with the equal-weight averaging direction from the eigenvalues that PR retains. A realizable equal-energy panel with nonnegative correlations raises PR by 14.3\% and squared error by 25\%. Empirically it is task-dependent within one pool: at $k\in\{4,8,16,24,31\}$, PR--error ranking agreement is 0.960--0.977 on MNLI-m but 0.223--0.419 on $\alpha$NLI, and scaling error by mean member energy raises the $\alpha$NLI figures to 0.937--0.960. The consensus direction retains 43.8\%, 33.7\%, and 35.9\% of centered residual variance on MNLI-m, SNLI, and $\alpha$NLI.
\item \textbf{Joint gains in accuracy and effective size exist and are reachable by at most two swaps.} Exhaustive enumeration at $k\in\{5,7\}$ on all four datasets shows that panels beating the accuracy-top-$k$ baseline on both accuracy and $\nu_H$ exist in all eight cases; the baseline never lies on the Pareto front. One bounded search, greedily swapping at most two members and inspecting under 2\% of the panels, raises $\nu_H$ by 24.8--56.0\% while raising accuracy by 0.10--1.10 percentage points. The models swapped in rank 15--32 in single-judge accuracy while those swapped out rank 1--6; the gain comes from removing mutually redundant strong models. The same panels improve distribution recovery in all eight cases and reach $96.0$--$100\%$ of the optimal $\nu_{\mathrm{MSE}}$ in the same candidate set under the same accuracy constraint ($70.6$--$100\%$ of the available gain).
\end{enumerate}

The public repository \citep{li2026votes} contains the final judge votes and analysis metadata for the ChaosNLI study and CC-1000. It supports audit and reuse without redistributing comment text or raw model responses. Equal member weights give majority labels for decisions and unthresholded label frequencies for distribution recovery. Sections~\ref{sec:related}--\ref{sec:experiments} position the contribution, define the measurements, and test the claims.

\section{Related Work}\label{sec:related}

\paragraph{Model judges and panels.}
Strong agreement with human preferences motivates model-based evaluation \citep{zheng2023}, but judge behavior can be unstable or exploitable \citep{thakur2025}. Multi-model panels can improve human agreement and reduce cost \citep{verga2024}. Cross-model consensus has also shown benefits over resampling one model for selecting reasoning answers \citep{liu2026}. We evaluate categorical inference and binary toxicity judgments against human label distributions, measuring both residual diversity and distributional recovery error.

\paragraph{Dependence and ensemble gains.}
Independence underlies classical voting and variance-reduction arguments \citep{condorcet1785,dietterich2000}. Correlated errors limit the gains from adding models \citep{kim2025,kohli2026b}, and cross-family similarity also appears in open-ended generation \citep{jiang2025}. Conditional-information criteria address member selection \citep{turkmen2026}; diversity audits distinguish member capability from diversity in majority-vote gains \citep{kim2026diversity}. Our distributional residuals retain label information lost by binary errors, and our recovery target evaluates the panel's full label frequencies. PR and an exact quadratic-form identity separate spectral diversity from recovery; the observed divergence depends on the task and tracks member energy.

\paragraph{Human disagreement and annotation budgets.}
Variation among human labels need not be removable noise \citep{pavlick2019,plank2022}. ChaosNLI provides dense item-level labels for studying such variation \citep{nie2020}. This differs from recovering a latent true label under annotator-error models \citep{dawid1979,raykar2010}: our target is a specified empirical distribution. Annotation allocation also depends on the objective. Stable ranking probabilities assess repeatability under labeling budgets \citep{riley2024}; sufficiently large budgets can favor labeling more items over relabeling the same items under particular binary-comparison assumptions \citep{dorner2026}. Soft-label learning can have different annotation saturation points for distribution matching and uncertainty ranking \citep{kohli2026a}. Prediction-powered inference formalizes how model predictions can supplement a smaller labeled sample for inference \citep{angelopoulos2023ppi}; Chaganty et al. quantify a related effective-data question for automatic metrics \citep{chaganty2018}. Our analysis holds model answers and items fixed and matches panel residuals to an explicit empirical human-distribution target.

\paragraph{Effective size and common variation.}
Design effects express variance inflation as an effective sample size \citep{kish1965}; in LLM judge panels, correlated errors can make a nominal panel yield only a small effective vote count \citep{kohli2026b}. Participation ratio measures spectral dimension \citep{mazzucato2016}. Human-label savings in assisted evaluation \citep{dorner2025} and cost-optimal allocation between judges \citep{angelopoulos2025} concern estimation precision and cost; $\nu_H$ concerns residual diversity. Common-direction analyses include market modes in correlation matrices \citep{laloux1999}, independent components under identification assumptions \citep{comon1994}, and constrained components informed by prior directions \citep{lu2005}. CARE separates quality and shared confounding under a latent-variable model \citep{zhao2026}; \citet{afrin2026} analyze common errors retained by aggregation and contamination from noisy references. These studies use different error targets. Our analysis retains the empirical human distribution, separates member energy from orientation relative to averaging, and uses a fixed equal-weight direction without independent-component identification.

\section{Human-Referenced Panel Measurements}\label{sec:measurements}
\subsection{Judgments and measurement targets}\label{sec:targets}
Let $n$ be the number of items, $k$ the number of judges, and $h_i\in\mathbb R^L$ the empirical human distribution over $L$ label categories on item $i$. Judge $a$ returns a categorical label $l(a,i)$, represented by a one-hot vector $e_{l(a,i)}$. The reference $h_i$ is an empirical target, not an error-free population distribution. We condition the measurements on this target and the fixed judge responses; the data and collection protocol are described in Section~\ref{sec:protocol}.

Table~\ref{tab:estimands} distinguishes the quantities used throughout the study and the question each one answers. The two human-reference sizes match different panel statistics. The variance share describes a component of residual averaging, while binary-error effective votes supply a comparison based on agreement with one gold label.
\begin{table}[t]
\centering\small
\setlength{\tabcolsep}{3pt}
\begin{tabularx}{\linewidth}{@{}l >{\raggedright\arraybackslash}X >{\raggedright\arraybackslash}X@{}}
\toprule
Quantity & Measurement target & Answers the question \\
\midrule
$\nu_H$ & PR of the normalized residual Gram matrix, matched to independent draws from $h$. & How many independent draws have comparable spectral diversity? \\
$\nu_{\mathrm{MSE}}$ & Squared error of the panel label distribution relative to $h$, matched to the same draws. & How many independent draws have comparable distributional error? \\
$\gamma_{\mathrm{co}}$ & Share of centered residual variance along the equal-weight judge direction. & How much centered variation survives equal-weight averaging? \\
$n_{\mathrm{eff}}$ & Signed-correlation summary of binary errors relative to the supplied gold label. & How many effective votes does the binary-error diagnostic report? \\
\bottomrule
\end{tabularx}
\caption{Measurement targets in the panel study.}\label{tab:estimands}
\end{table}

\subsection{Residual representation}\label{sec:residual}
Judge $a$'s residual relative to the empirical human label distribution is
\begin{equation}\label{eq:residual}
r_{a,i}=e_{l(a,i)}-h_i\in\mathbb R^L.
\end{equation}
Residual coordinates sum to zero. We use an orthonormal basis of this $(L-1)$-dimensional subspace: for the ordered labels entailment, neutral, contradiction, $v_1=(1,0,-1)^\top/\sqrt2$ and $v_2=(-1,2,-1)^\top/\sqrt6$; for binary labels, $v_1=(1,-1)^\top/\sqrt2$. The two three-label coordinates preserve interpretable axes: $v_1$ measures polarity and $v_2$ measures neutrality excess relative to the extremes.

Let $c_{a,i}$ contain the coordinates $v_t^\top r_{a,i}$. Averaging $c_{a,i}c_{b,i}^\top$ over items gives a second-moment block $B_{a,b}$. We form a normalized residual Gram matrix
\begin{equation}\label{eq:residual_gram}
C_{a,b}=\frac{\operatorname{tr}B_{a,b}}
{\sqrt{\operatorname{tr}B_{a,a}\operatorname{tr}B_{b,b}}}.
\end{equation}
This is the normalized inner product between flattened residual arrays. We retain their across-item means, so $C$ includes average displacement from the reference. With positive residual energy for each judge, $C$ is positive semidefinite, has unit diagonal, and has $\operatorname{tr}(C)=k$. Taking block traces discards the full cross-label block structure.

\subsection{Spectral effective size \texorpdfstring{$\nu_H$}{nuH}}\label{sec:spectral_size}
The participation ratio (PR) measures effective spectral dimension \citep{mazzucato2016}. For eigenvalues $\lambda_j$ of $C$ and the mean squared inner product $\bar q$ over unordered off-diagonal pairs, its unit diagonal gives
\begin{equation}\label{eq:pr}
\mathrm{PR}
=\frac{(\sum_{j=1}^k\lambda_j)^2}{\sum_{j=1}^k\lambda_j^2}
=\frac{k}{1+(k-1)\bar q}.
\end{equation}
PR lies in $[1,k]$; it measures spectral concentration and does not retain inner-product signs. For $k=1$, we set $\mathrm{PR}=1$.

We express PR in human-reference units using direct Monte Carlo sampling \citep{robert2004}. Conditional on $h$, draw each of $m$ simulated labels per item independently from $\operatorname{Cat}(h_i)$ and compute the resulting panel PR. At each grid size, the mean of $N=12$ replicates is $\mathrm{PR}_0(m)$. This fixed computational budget and its replicate spread are documented in Appendix~\ref{app:protocol}. For the first adjacent grid points that bracket $\mathrm{PR}_{\mathrm{obs}}$, define
\begin{equation}\label{eq:human_size}
\begin{aligned}
\nu_H={}&m_j+(m_{j+1}-m_j)\\
&\quad\cdot\frac{\mathrm{PR}_{\mathrm{obs}}-\mathrm{PR}_0(m_j)}
{\mathrm{PR}_0(m_{j+1})-\mathrm{PR}_0(m_j)}.
\end{aligned}
\end{equation}
The grid is $2,\ldots,12,16,24,32,48,64,96,128$. All mean reference curves used here are strictly increasing. Under this Monte-Carlo curve, values at or below the first mean are reported as $\nu_H\leq2$, and values above the final mean are out of range.

We call $\nu_H$ the \emph{human-referenced effective panel size}. It may be noninteger and differs from the nominal panel size $k$. Within one strictly increasing reference curve, it preserves PR's ranking. This is exact under the analytic calibration below and holds above the censored lower end under the Monte-Carlo curve. The matching supplies units without adding ranking information. Its scale depends on $h$, the item set, and the residual and simulation protocols. At finite $n$, $\mathrm{PR}_0(m)$ is generally below $m$. For fixed $m$, independent item draws and a positive lower bound on average human-label variance suffice for $\mathrm{PR}_0(m)\to m$ as $n$ grows. $\nu_H$ matches spectral diversity; distributional error is matched on its own target in Section~\ref{sec:recovery}.

\paragraph{An analytic approximation to the reference.}
Panel selection requires calibration below the grid's $m=2$ boundary. We derive a closed-form approximation for these comparisons. For two conditionally independent draws from $h$ over $n$ items, replacing their squared inner product and their two residual energies by expectations gives
\begin{equation}\label{eq:delta_closed}
\begin{aligned}
\delta&=\frac{\langle s_2-2s_3+s_2^2\rangle_i}{n\langle1-s_2\rangle_i^2},\\[-1pt]
s_2&=\sum_lh_{il}^2,\qquad s_3=\sum_lh_{il}^3,
\end{aligned}
\end{equation}
which has no free parameter. It is a ratio of expectations rather than the expected squared normalized inner product, and the two differ at small $n$. Substituting $\bar q=\delta$ in Equation~\eqref{eq:pr} gives the approximate reference $\mathrm{PR}_\delta(m)=m/(1+(m-1)\delta)$, whose exact inverse is
\begin{equation}\label{eq:nu_closed}
\nu_H=\frac{\mathrm{PR}(1-\delta)}{1-\mathrm{PR}\,\delta}.
\end{equation}
Equation~\eqref{eq:nu_closed} is defined and strictly increasing on $\mathrm{PR}\in[1,1/\delta)$, so it requires no censoring and preserves the PR order of every candidate panel exactly, including below two draws. Our item sets give $\delta=7.80\times10^{-4}$, $1.02\times10^{-3}$, $1.93\times10^{-3}$, and $1.15\times10^{-3}$ on MNLI-m, SNLI, $\alpha$NLI, and CC-1000, so $1/\delta>517$ and every panel we report lies inside that domain. At $n=1000$, this approximation agrees closely with Monte Carlo: over every grid point on all four item sets the largest gap between $\mathrm{PR}_\delta(m)$ and $\mathrm{PR}_0(m)$ is $0.18\%$, and the resulting $\nu_H$ differ by $0.027\%$--$0.077\%$ on the four full 32-judge panels, with a largest difference of $0.15\%$ on the selected panels the grid covers. All panel-selection $\nu_H$ values use Equation~\eqref{eq:nu_closed}; other $\nu_H$ values use the Monte-Carlo curve of Equation~\eqref{eq:human_size}.

\subsection{What spectral size implies for distribution recovery}\label{sec:recovery}
To compare residual diversity with distribution recovery, we measure the squared error of the panel's label frequencies. Let $\bar p_i=k^{-1}\sum_a e_{l(a,i)}$ be its label frequencies, with no majority-vote threshold. Define
\begin{equation}\label{eq:loss_E}
\mathcal E=\frac1n\sum_i\|\bar p_i-h_i\|^2.
\end{equation}
For $m$ conditionally independent human-reference labels per item, the expected error is $J/m$, where $J=n^{-1}\sum_i(1-\|h_i\|^2)$. Matching this target yields
\begin{equation}\label{eq:loss_human}
\nu_{\mathrm{MSE}}=\frac{J}{\mathcal E}.
\end{equation}
It is finite and positive when $J,\mathcal E>0$; no finite match exists for $\mathcal E=0<J$, and the reference degenerates when $J=0$. It is the intersection of panel error with the human-draw curve $J/m$. In the fixed-moment extension of Section~\ref{sec:effective_sizes}, panel error reaches a positive floor and $\nu_{\mathrm{MSE}}$ saturates.

Let $\langle\cdot\rangle_i$ denote an item average, $K_{a,b}=\langle r_{a,i}^\top r_{b,i}\rangle_i$ the unnormalized Gram matrix, $\omega_a=K_{a,a}$ the member energy, and $D=\operatorname{diag}(\sqrt{\omega_1},\ldots,\sqrt{\omega_k})$. Then
\begin{equation}\label{eq:loss_gram}
C=D^{-1}KD^{-1},\qquad
\mathcal E=\frac1{k^2}\mathbf1^\top K\mathbf1.
\end{equation}
PR uses the eigenvalues of normalized $C$, whereas error uses a quadratic form of $K$. The following identity makes this distinction explicit.

\par\vspace*{8pt}
\noindent\begin{minipage}{\linewidth}
\noindent\textbf{Exact identity (error and spectral orientation).}\enspace
Assume $\omega_a>0$ for all members, so the normalized Gram matrix is defined. Write $C=\sum_{j=1}^k\lambda_jw_jw_j^\top$ with orthonormal $w_j$, and let $d=(\sqrt{\omega_a})_{a=1}^k$, $\Omega=d^\top d$, and $b_j=(w_j^\top d)^2/\Omega$. Then
\begin{equation}\label{eq:loss_orientation}
\mathcal E=\frac{\Omega}{k^2}\sum_{j=1}^k\lambda_jb_j,
\qquad \sum_{j=1}^k b_j=1.
\end{equation}
\end{minipage}
\par\vspace*{8pt}
\noindent Here $b_j$ is the squared alignment of the energy-weighted averaging direction with eigenvector $w_j$. Error depends on these weights and the energy scale $\Omega$, while PR depends only on $\sum_j\lambda_j^2$. The identity follows by substituting the eigendecomposition into $d^\top C d/k^2$; Appendix~\ref{app:loss} discusses repeated eigenvalues and gives realizable equal-spectrum panels with different errors. Normalizing members removes their energy scales, and retaining eigenvalues alone discards orientation.

Even equal member energies do not make PR an error ranking. If $\omega_a=\omega$, and $\bar c_C,v_C$ are the mean and variance of off-diagonal $C_{a,b}$, then
\begin{equation}\label{eq:loss_equal_energy}
\bar q=\bar c_C^{\,2}+v_C,\qquad
\frac{\mathcal E}{\omega}=\frac{1+(k-1)\bar c_C}{k}.
\end{equation}
The second moment $\bar q$ controls PR, while the signed first moment controls error. Their different dependence on $v_C$ permits conflicting changes. A realizable construction with $k=4$ and $h_i=(1/2,1/2)$ has equal energies, zero mean residuals, and nonnegative correlations, yet increasing PR from $2$ to $16/7$ increases $\mathcal E$ from $1/4$ to $5/16$. Appendix~\ref{app:loss} gives the labels and derivations. Larger PR or $\nu_H$ therefore does not guarantee smaller $\mathcal E$ or larger $\nu_{\mathrm{MSE}}$, even in this benign setting. Section~\ref{sec:spectral_vs_recovery} examines the empirical extent of this mismatch.

\subsection{Consensus-direction variance share}\label{sec:consensus_share}
To measure variation retained by equal-weight averaging, we project onto the fixed judge-space direction $u_1=\mathbf1/\sqrt{k}$ \citep{afrin2026}. For label coordinate $t$, form $X_{i,a}=(c_{a,i})_t$, center each column across items to obtain $X_c$, and let $S=X_c^\top X_c/n$.

For $k\geq2$ and $\operatorname{tr}S>0$, write $\bar v=\operatorname{tr}S/k$ for mean judge variance and $\bar c$ for mean off-diagonal covariance. Their ratio is
\begin{equation}\label{eq:rho_bar}
\bar\rho=\frac{\bar c}{\bar v}
=\frac{\sum_{a\ne b}S_{a,b}}{(k-1)\operatorname{tr}S},
\qquad \bar v=\frac{\operatorname{tr}S}{k}.
\end{equation}
This normalized mean covariance generally differs from the mean of pairwise Pearson correlations. The consensus-direction variance share is
\begin{equation}\label{eq:consensus}
\gamma_{\mathrm{co}}
=\frac{u_1^\top S u_1}{\operatorname{tr}S}
=\frac{1+(k-1)\bar\rho}{k}.
\end{equation}
It lies in $[0,1]$ and equals $1/k$ when mean covariance is zero, linking it to the design-effect form \citep{kish1965}. It measures centered shared fluctuations, not statistical bias or the across-item mean residual. Across label coordinates, we use total-variance weighting:
$\gamma_{\mathrm{co,all}}=\sum_tu_1^\top S_tu_1/\sum_t\operatorname{tr}S_t$, with positive total trace. Unless a label coordinate is specified, we use $\gamma_{\mathrm{co}}$ as shorthand for this pooled share.

Only the consensus component survives the full equal-weight average; components orthogonal to $\mathbf1$ cancel. For a uniformly selected subset $I$ of $m$ judges from this fixed pool, let $\bar x_I=m^{-1}\sum_{a\in I}(X_c)_{:,a}$. Using empirical variance with denominator $n$,
\begin{equation}\label{eq:subset_variance}
\begin{aligned}
V_m&=\mathbb E_{I:|I|=m}[\operatorname{Var}_n(\bar x_I)]\\
&=\bar c+\frac{\bar v-\bar c}{m},\qquad 1\leq m\leq k.
\end{aligned}
\end{equation}
Thus $V_m/\bar v=1/m+(1-1/m)\bar\rho$. Substituting the full panel's observed $\bar\rho$ gives an analytic curve for averaging within the fixed pool.

The consensus share connects centered variation to the distributional error defined in Section~\ref{sec:recovery}. For $\mu_a=\langle r_{a,i}\rangle_i$ and $\bar\mu=k^{-1}\sum_a\mu_a$, the raw-scale channel covariances give
\begin{equation}\label{eq:loss_consensus}
\mathcal E=\|\bar\mu\|^2+
\frac{\sum_t\operatorname{tr}S_t}{k}\gamma_{\mathrm{co,all}}.
\end{equation}
The two terms are the squared mean residual and the variation of the aggregated residual across items. The share $\gamma_{\mathrm{co,all}}$ must therefore be interpreted together with total variance and mean displacement \citep{afrin2026}.

\subsection{Strict joint improvement over the top-\texorpdfstring{$k$}{k} baseline}\label{sec:panel_search}
The fixed-pool search asks whether a panel can improve accuracy and spectral size simultaneously. For a panel $S$ of fixed size $k$, let $\mathrm{acc}(S)$ be the agreement between its majority vote and $\mathrm{gold}_i$, scoring ties as the expectation under uniform tie-breaking rather than resolving them. Equation~\eqref{eq:pr} makes the search tractable: PR depends on the panel only through the mean squared inner product $\bar q(S)$, and the analytic calibration of Equation~\eqref{eq:nu_closed} is strictly increasing in PR, so $\nu_H$ is a strictly decreasing function of $\bar q$ with no censoring at the low end. Since $\bar q$ is determined by the pairwise entries of $C$, any candidate panel's $\nu_H$ reads off a precomputed $C$ with no further simulation.

The baseline $S_0$ consists of the $k$ most accurate available models with equal weights. We define the joint-improvement set
\begin{equation}\label{eq:joint_improvement}
\begin{aligned}
D(S_0)=\{S:\ &\mathrm{acc}(S)>\mathrm{acc}(S_0),\\[-1pt]
             &\nu_H(S)>\nu_H(S_0)\},
\end{aligned}
\end{equation}
with both inequalities strict and evaluated on the same items. Exhaustive enumeration in Section~\ref{sec:panel_search_eval} finds $D(S_0)\neq\varnothing$ in all eight dataset--size cases.

A single greedy step supplies a feasible path: choose the best panel in a bounded neighborhood of $S_0$. Replacing one or two members of $S_0$ gives the candidate sets $N_1$ and $N_2$, with $k(32-k)$ and $\binom{k}{2}\binom{32-k}{2}$ panels. On $N=N_1\cup N_2$, with the feasible set $N^+=\{S\in N:\mathrm{acc}(S)>\mathrm{acc}(S_0)\}$, we report three rules:
\begin{description}\setlength{\itemsep}{1pt}\setlength{\parsep}{0pt}
\item[Rule A ($\max\nu_H$).] $S_A=\arg\max_{S\in N^+}\nu_H(S)$: accuracy is the constraint and effective size the objective.
\item[Rule D ($\max\mathrm{acc}$).] $S_D=\arg\max_{S\in N}\mathrm{acc}(S)$, with accuracy ties broken by $\nu_H$.
\item[Rule E ($\min\mathcal E$).] $S_E=\arg\min_{S\in N^+}\mathcal E(S)$, with $\mathcal E$ the distributional error of Section~\ref{sec:recovery}; the direct-recovery control.
\end{description}
Remaining ties go to the first candidate in the enumeration order of Appendix~\ref{app:selection_protocol}. Because PR discards member energy and orientation, a panel chosen for spectral size need not recover the distribution well; rule E selects for recovery directly, so comparing the two answers what selecting on $\nu_H$ costs on the target it does not optimize.

\section{Experiments}\label{sec:experiments}
\subsection{Data and audit protocol}\label{sec:protocol}
We use MNLI-m, SNLI, and $\alpha$NLI from ChaosNLI \citep{nie2020}. MNLI-m and SNLI distinguish entailment, neutral, and contradiction \citep{williams2018,bowman2015}; $\alpha$NLI chooses between two abductive hypotheses \citep{bhagavatula2020}. Each selected ChaosNLI item has 100 human labels. We sample 1,000 items per dataset approximately equally across human-entropy terciles, with seed 42. A full-population recheck with six replacement judges changes $\nu_H$ by up to 1.3\% and $\nu_{\mathrm{MSE}}$ by up to 5.6\% relative to its retained sample. It confirms joint selection gains on all three full tasks (Appendix~\ref{app:population}). Task agreement uses the dataset-provided 100-label mode $\mathrm{gold}_i$, preserving its original tie choices. We additionally check the same measurements on one external set, CC-1000: a frozen 1,000-item subset drawn from the Civil Comments toxicity data \citep{borkan2019,jigsaw2019}.

For the three ChaosNLI tasks, the fixed panel contains 32 requested model identifiers (Appendix~\ref{app:protocol}, Table~\ref{tab:judge_panel}). Each ChaosNLI model responds separately at temperature 0, with one user message, no system message, and a label-only instruction. Requested model identifiers are recorded in the supplement; they do not independently authenticate the service's underlying provider version. Table~\ref{tab:protocol} summarizes the collection and analysis design.

\begin{table}[t]
\centering\small
\setlength{\tabcolsep}{3pt}
\begin{tabularx}{\linewidth}{@{}l >{\raggedright\arraybackslash}X@{}}
\toprule
Component & Scope and treatment \\
\midrule
Baseline & $32\times1{,}000\times3=96{,}000$ records; 95,994 parseable. The 6 failures (0.00625\%) retain flagged placeholders: 1/0/5 by task. \\
Human reference & 100 labels per item; $h_i$ is their empirical frequency vector. \\
Panel audit & 54,139 unique member sets per task; 150,372 fixed member additions. Shared judges and items make these overlapping comparisons. \\
Item stability & Fixed 500/500 halves; compares stability within observed data. Failed-item sensitivity is recorded separately. \\
Presentation & 122,000 records, including cache reuse; its own original-response snapshot. Complete-case sets contain 489/492/492 items and 31/31/29 judges. \\
CC-1000 check & 1,000 items, 32 judges, and 176,251 human toxicity votes; score-only public release. \\
\bottomrule
\end{tabularx}
\caption{Collection and analysis design.}\label{tab:protocol}
\end{table}

The baseline results use the recorded matrices. Six deterministic placeholder labels are treated as invalid responses (1 on MNLI-m, 0 on SNLI, and 5 on $\alpha$NLI) and excluded from the model-judgment count. Excluding the affected items changes the full-panel distribution error by less than 0.3\% and leaves the stable-conflict pattern intact (Appendix~\ref{app:failure_sensitivity}). The label, tie, filtering, and reference-draw rules are in Appendix~\ref{app:protocol}; the supplementary reproduction map links each result to its available records and saved outputs.

\subsection{Effective sizes and reference choices}\label{sec:effective_sizes}
On the three ChaosNLI tasks, distributional-error matching gives $\nu_{\mathrm{MSE}}=2.304$, $3.750$, and $3.445$, while spectral matching gives $\nu_H=4.242$, $6.459$, and $6.499$ (Table~\ref{tab:effective_sizes}). The two targets therefore differ by factors of 1.841, 1.723, and 1.887; the binary-error diagnostic remains 1.971, 2.227, and 1.999 effective votes. The gap persists when human disagreement is retained in the residual reference.

\begin{table}[t]
\centering\small
\setlength{\tabcolsep}{4pt}
\begin{tabular}{@{}lrrr@{}}
\toprule
Measurement & MNLI-m & SNLI & $\alpha$NLI \\
\midrule
$n_{\mathrm{eff}}$ & 1.97 & 2.23 & 2.00 \\
$\nu_H$ & 4.24 & 6.46 & 6.50 \\
$\nu_{\mathrm{MSE}}$ & 2.30 & 3.75 & 3.44 \\
$\nu_H/\nu_{\mathrm{MSE}}$ & 1.84 & 1.72 & 1.89 \\
$\mathcal E$ & 0.19691 & 0.09068 & 0.04825 \\
$\gamma_{\mathrm{co,all}}$ & 43.8\% & 33.7\% & 35.9\% \\
\bottomrule
\end{tabular}
\caption{Measurement values for the same 32-judge panels on ChaosNLI.}\label{tab:effective_sizes}
\end{table}

The same decomposition gives a fixed-pool asymptote for the distributional target. Let $\mu_2=\|\bar\mu\|^2$, $\bar v= (\mathcal E-\mu_2)/\gamma_{\mathrm{co,all}}$, and $\bar\rho=(k\gamma_{\mathrm{co,all}}-1)/(k-1)$ at $k=32$. Under an extension that holds the squared mean residual $\mu_2$, mean member variance $\bar v$, and normalized mean centered covariance $\bar\rho$ fixed for additional members,
\begin{equation}\label{eq:pool_ceiling}
\mathcal E_\infty=\mu_2+\bar v\bar\rho,
\qquad \nu_{\mathrm{MSE},\infty}=J/\mathcal E_\infty.
\end{equation}
For MNLI-m, SNLI, and $\alpha$NLI this gives $\nu_{\mathrm{MSE},\infty}=2.392$, $3.990$, and $3.655$; the observed 32-judge values are 96.3\%, 94.0\%, and 94.2\% of them. In the same curve, projected values at $k=64$ are only 0.043, 0.117, and 0.102 higher than at $k=32$. Which moments are held fixed matters for where the asymptote sits. Holding the \emph{uncentered} Gram moments fixed instead, with mean member energy $\bar\omega$ and mean off-diagonal $\bar c_K$, makes $\mathbb E_{|I|=m}[\mathcal E(I)]=\bar c_K+(\bar\omega-\bar c_K)/m$ an exact fixed-pool curve whose limit is $\bar c_K$, which differs from $\mathcal E_\infty$ by $k^{-1}\sum_a\|\mu_a-\bar\mu\|^2/(k-1)$ and gives $2.397$, $4.003$, and $3.656$ (Appendix~\ref{app:loss}). The difference is under $0.4\%$, so the diminishing-returns reading holds under either extension. Section~\ref{sec:panel_search_eval} quantifies the gains from choosing members.

Binary-error effective votes $n_{\mathrm{eff}}$ are approximately 2 \citep{kish1965,kohli2026b}. The difference from $\nu_H$ involves both representation, including encoding, reference, and centering, and summarization. Keeping the binary-error Pearson matrix but replacing the signed-correlation summary by PR gives 3.54, 4.38, and 3.79. Using human-distribution residuals then gives PR of 4.23, 6.42, and 6.43. The conversion from this PR to $\nu_H$ changes the values by only 0.3\%--1.1\%; the complete representation and summarization account for most of the difference from binary effective votes. Table~\ref{tab:anchor_comparison} isolates the anchor, varying it within the same uncentered residual construction. Appendix~\ref{app:loss} gives the crossed comparison.

The next comparison keeps the answers and spectral summary fixed while changing the residual anchor $A$: uniform labels, a human-mode vector $A_{\mathrm{mode}}$, or the full distribution $h$. Ties in $A_{\mathrm{mode}}$ take the first label in the order of Section~\ref{sec:residual}; this can differ from the supplied $\mathrm{gold}_i$. Simulated labels always come from $h$, while both model and simulated residuals subtract the same $A$, with a separate reference curve for each anchor.

\begin{table}[t]
\centering\small
\setlength{\tabcolsep}{4pt}
\begin{tabular}{@{}lrrr@{}}
\toprule
Anchor $A$ & MNLI-m & SNLI & $\alpha$NLI \\
\midrule
\multicolumn{4}{c}{Mean squared inner product $\bar q$} \\
\midrule
Uniform $1/L$ & 0.412 & 0.531 & 0.676 \\
Mode $A_{\mathrm{mode}}$ & 0.425 & 0.284 & 0.287 \\
Distribution $h$ & 0.212 & 0.128 & 0.128 \\
\midrule
\multicolumn{4}{c}{Effective size $\nu_H$} \\
\midrule
Uniform $1/L$ & 2.75 & 2.57 & 2.33 \\
Mode $A_{\mathrm{mode}}$ & 2.76 & 4.32 & 3.94 \\
Distribution $h$ & 4.24 & 6.46 & 6.50 \\
\bottomrule
\end{tabular}
\caption{Effective sizes under three residual anchors.}\label{tab:anchor_comparison}
\end{table}

Using the full distribution as the residual anchor yields lower $\bar q$ and higher $\nu_H$ on every task (Table~\ref{tab:anchor_comparison}). The mode and uniform anchors give similar effective sizes on MNLI-m; using the mode already raises the value on SNLI and $\alpha$NLI. Retaining the full distribution raises it further on all three. The anchor therefore changes which dependence is visible in the same model answers.

\subsection{Spectral diversity versus distribution recovery}\label{sec:spectral_vs_recovery}
Section~\ref{sec:recovery} rules out a general monotone relationship between $\nu_H$ and distribution recovery. In the observed pool, ranking agreement depends on the task, and member additions can move the two objectives in opposite directions.

\subsubsection{Within-size rankings and member energy}
For each $k=2,\ldots,31$, we deduplicate the 2,000 sampled panels by member set and compute the Spearman correlation of PR with $-\mathcal E$. The minus sign makes larger values favorable for both quantities. Within a strictly increasing reference curve's matched range, these rankings also correspond to those of $\nu_H$ and $\nu_{\mathrm{MSE}}$. Working with PR retains small panels below the human-matching lower limit.

Because normalization removes member energy, we also compare PR with $-\mathcal E/\bar\omega$, where $\bar\omega=k^{-1}\sum_a\omega_a$ is mean member residual energy. This measures error relative to the members' average squared deviation.

\begin{figure}[t]
\centering
\includegraphics[width=\linewidth]{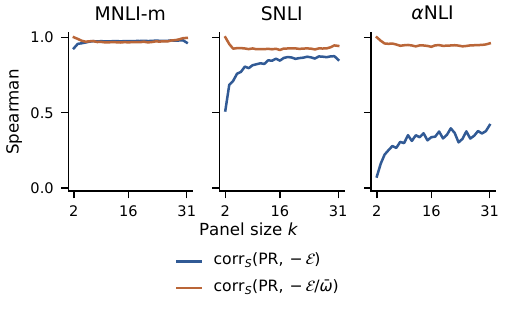}
\caption{Within-size Spearman correlations between spectral and distribution-recovery rankings.}\label{fig:loss_association}
\end{figure}

At sizes $k=4,8,16,24,31$, the correlation between PR and $-\mathcal E$ ranges from 0.960 to 0.977 on MNLI-m and from 0.711 to 0.867 on SNLI, but only from 0.223 to 0.419 on $\alpha$NLI (Figure~\ref{fig:loss_association}). Spectral and distributional rankings align closely on MNLI-m and diverge sharply on $\alpha$NLI.

Scaling error by member energy raises the $\alpha$NLI correlations to 0.937--0.960; at $k=16$, the increase is from 0.338 to 0.937. The ranking mismatch is closely associated with member error energy: a panel can offer diverse residual directions while its members deviate farther from the human distribution. This motivates reporting member error energy together with diversity \citep{kim2026diversity}.

\subsubsection{Member additions and item stability}
For each unique panel at $k=4,8,16,24,31$, we add every model outside the panel. Define spectral gain $\Delta\mathrm{PR}=\mathrm{PR}_{\mathrm{new}}-\mathrm{PR}_{\mathrm{old}}$ and error improvement $\Delta\mathcal E=\mathcal E_{\mathrm{old}}-\mathcal E_{\mathrm{new}}$. Opposite signs mean the objectives conflict.

We repeat the calculation on two fixed 500-item halves. Table~\ref{tab:loss_additions} counts additions with the same direction of conflict on the full set and both halves, and with both relative changes at least 1\% on each set. The relative changes use the original panel's values: $|\Delta\mathrm{PR}|/\mathrm{PR}_{\mathrm{old}}$ and $|\Delta\mathcal E|/\mathcal E_{\mathrm{old}}$. We use a fixed descriptive 1\% filter; the split assesses stability within the observed items.

\begin{table}[!htbp]
\centering\small
\setlength{\tabcolsep}{3pt}
\begin{tabular}{@{}crrrr@{}}
\toprule
Addition & Examined & MNLI-m & SNLI & $\alpha$NLI \\
\midrule
$4\to5$ & 54,348 & 2,116 & 309 & 1,127 \\
$8\to9$ & 48,000 & 332 & 441 & 1,085 \\
$16\to17$ & 32,000 & 0 & 7 & 447 \\
$24\to25$ & 15,992 & 0 & 0 & 85 \\
$31\to32$ & 32 & 0 & 0 & 0 \\
\bottomrule
\end{tabular}
\caption{Stable conflicting member additions by panel size.}\label{tab:loss_additions}
\end{table}

For $4\to5$, 3.89\%, 0.57\%, and 2.07\% of additions meet this criterion. A member can improve spectral diversity while worsening distribution recovery, or the reverse. The frequencies generally decline with size; none of the $31\to32$ additions passes the filter.

Panel rankings themselves depend on item composition. The correlation of PR rankings between the two halves is only 0.370--0.554 on $\alpha$NLI, compared with 0.915--0.942 on MNLI-m. Appendix~\ref{app:diagnostics} gives the full curves and protocol. The within-size associations and stable conflicts answer different questions. A high overall correlation can coexist with individual reversals. Distributional error evaluates the recovery target directly; spectral diversity measures a different property of the same responses.

\subsection{Geometry of model and constructed residuals}\label{sec:geometry}
\begin{figure}[tbp]
\centering
\includegraphics[width=0.84\linewidth]{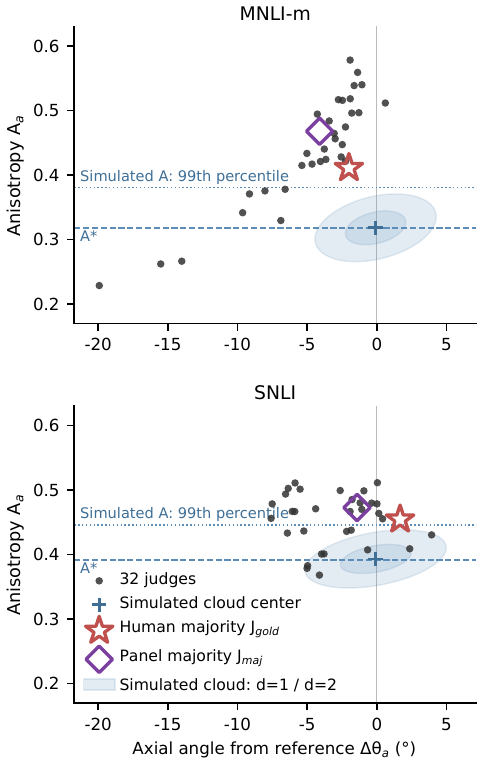}
\caption{Residual geometry relative to human sampling.}\label{fig:geometry}
\end{figure}

Model residuals differ from human sampling noise in both shape and orientation. Figure~\ref{fig:geometry} compares the two on the three-label tasks. For each judge, anisotropy $A_a$ is the difference between its two residual second-moment eigenvalues divided by their sum; larger values mean greater concentration along one axis. The angle $\Delta\theta_a$ measures rotation of that principal axis from the analytic human reference. Simulated labels drawn itemwise from $h$ give the comparison cloud. Appendix~\ref{app:geometry} defines these coordinates and cloud distances.

Most model points lie above and to the left of the simulated cloud center: 28 of 32 on MNLI-m and 23 of 32 on SNLI. Upward displacement means greater anisotropy; leftward displacement means a negative principal-axis rotation. The constructed judge $J_{\mathrm{gold}}$, which returns $\mathrm{gold}_i$ without model inference, also lies above the cloud. Its anisotropy excess is 76.9\% and 99.1\% of the mean model excess. Both this judge and the simulations output one-hot labels, but one deterministically chooses the human mode while the other samples from $h$. Thus mode selection can produce similar upward displacement; one-hot encoding alone does not explain the contrast.

Aggregating all 32 judges into $J_{\mathrm{maj}}$ shifts the point left relative to $J_{\mathrm{gold}}$ by approximately $2.10^\circ$ and $3.09^\circ$. Here $J_{\mathrm{maj}}$ returns the panel's modal label, resolving ties in canonical label order. Majority vote retains a directional deviation shared by the individual model points. The next section measures the judge-space direction whose variation survives equal-weight averaging.

\subsection{Shared variation retained by averaging}\label{sec:shared_variation}

Using all 32 judges, the two channel-specific consensus-direction variance shares $\gamma_{\mathrm{co}}$ are 44.8\% and 43.4\% on MNLI-m, and 36.0\% and 32.8\% on SNLI. Total-variance weighting gives 43.8\% and 33.7\%; the single $\alpha$NLI channel gives 35.9\% (Table~\ref{tab:effective_sizes}). These shares quantify centered residual variation retained along the fixed equal-weight direction.

\begin{figure}[tbp]
\centering
\includegraphics[width=0.88\linewidth]{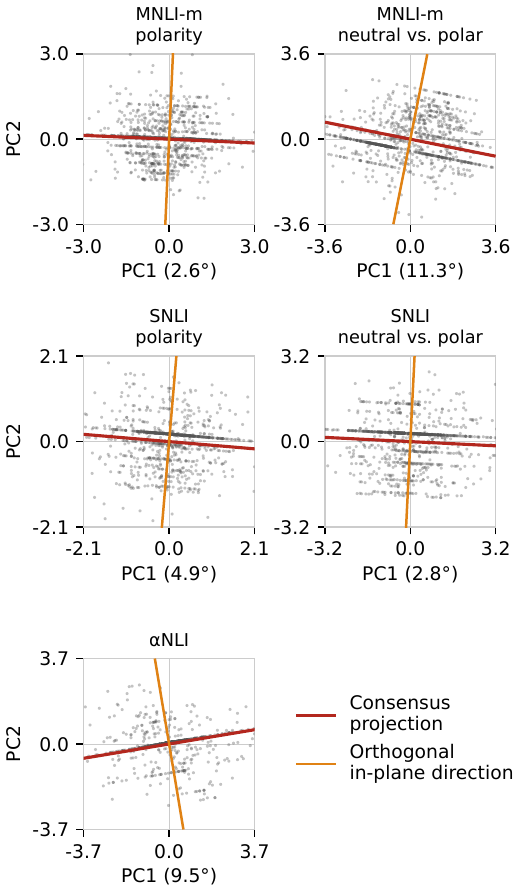}
\caption{Projection of consensus and orthogonal directions in a 10-model panel.}\label{fig:consensus_projection}
\end{figure}

Figure~\ref{fig:consensus_projection} illustrates the same alignment in a separate 10-model panel with one member per family. The projection makes the consensus direction and residual geometry visible in two principal components; the full consensus vector need not lie in that plane.

\begin{figure}[tbp]
\centering
\includegraphics[width=\linewidth]{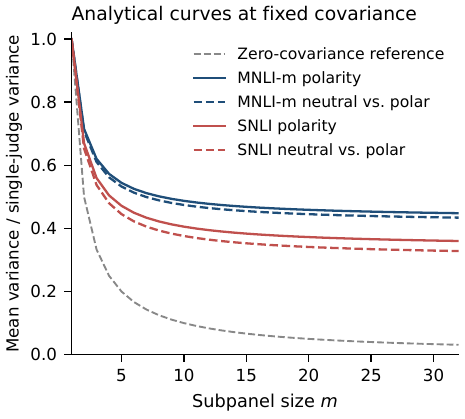}
\caption{Expected variance of a random-subpanel mean.}\label{fig:subpanel_variance}
\end{figure}

Equations~\eqref{eq:consensus}--\eqref{eq:subset_variance} show how this common variation survives averaging and produces diminishing variance reduction (Figure~\ref{fig:subpanel_variance}). Decomposing full-panel distributional error with Equation~\eqref{eq:loss_consensus}, the squared mean residual $\|\bar\mu\|^2$ accounts for 11.3\%, 4.8\%, and 0.1\% of $\mathcal E$. Most error thus comes from aggregated residuals that vary across items, rather than a fixed across-item displacement.

\subsection{Diminishing returns from panel expansion}\label{sec:panel_size}
\begin{figure}[tbp]
\centering
\includegraphics[width=\linewidth]{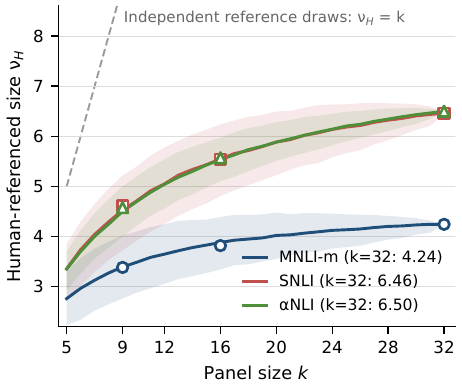}
\caption{Random-subpanel $\nu_H$ by panel size.}\label{fig:scale_curve}
\end{figure}

The median $\nu_H$ of random subpanels increases with nominal size, with diminishing gains (Figure~\ref{fig:scale_curve}).

Searched subpanels exceed the full panel's $\nu_H$, while provider-family diversity yields task-dependent gains (Appendix~\ref{app:diagnostics}). Option presentation provides another source of residual diversity.

\subsection{Presentation changes and residual association}\label{sec:presentation}
Changing how the same options are presented does move judges' labels, but the moved labels retain more residual association than matched independent transitions. We replace one judge at a time by its option-reordered responses, keeping all other responses fixed. Define $\Delta\bar q=\bar q_{\mathrm{alt}}-\bar q_{\mathrm{base}}$, so negative values indicate reduced squared residual association. An independent-transition reference samples each item's replacement from that judge's empirical transition probabilities conditional on its original label. This reference includes unchanged labels and matches conditional flip rates and destination frequencies in expectation.

Missing variants leave $31+31+29=91$ model--dataset combinations, using 489, 492, and 492 common complete-case items. The comparisons use the presentation experiment's own stored original-response panel, with cache reuse in the response records. They cannot isolate presentation-order effects from repeat-call variation. Appendix~\ref{app:protocol} gives the full response and simulation protocols.

\begin{figure}[tbp]
\centering
\includegraphics[width=0.84\linewidth]{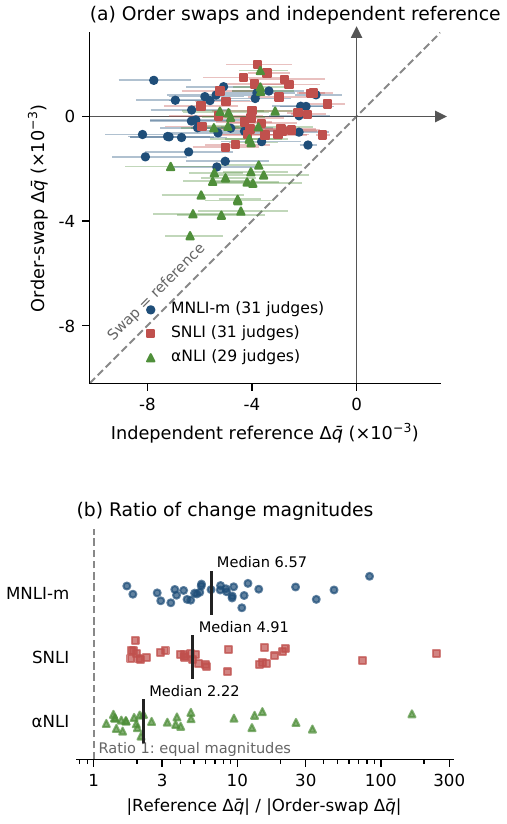}
\caption{Observed presentation changes and an independent-transition reference.}\label{fig:presentation}
\end{figure}

Under the independent-transition reference, points would fluctuate around $y=x$ in Figure~\ref{fig:presentation}a. Instead, 86 of 91 descriptive comparisons lie above their corresponding reference's 97.5th percentile; the other 5 fall within its interval. The reordered responses retain more residual association than the matched independent-transition draws. The reference mean change is negative for every combination, whereas the observed change is positive in 37 cases. The median absolute-change ratios are 6.57, 4.91, and 2.22 (Figure~\ref{fig:presentation}b). The observed response changes therefore differ from independent transitions with the same conditional flip rates. Additional flip and subpanel diagnostics appear in Appendix~\ref{app:diagnostics}.

\subsection{Sensitivity to the human reference}\label{sec:reference_sensitivity}
How many human labels define the reference also matters: the measured effective size approaches its full-reference value as that number grows. Keeping all 32 model judges fixed, we estimate an anchor $A_M$ by drawing $M$ labels without replacement from each item's 100-label count pool. Model and simulated residuals subtract $A_M$, while simulated labels still come from the full $h$. This changes reference estimation while holding model answers fixed.

Median $\nu_H$ is 3.35, 4.36, and 4.32 at $M=5$, and 3.70, 5.10, and 5.12 at $M=10$. It approaches the full-reference values 4.24, 6.46, and 6.50 as $M$ increases; distributions at adjacent larger $M$ overlap substantially. Appendix~\ref{app:protocol}, Figure~\ref{fig:anchor_humans}, gives the complete sensitivity curves.

Anchor error enters every model residual as the same vector,
$r_{a,i}^{A}=r_{a,i}^{h}+(h_i-A_{M,i})$.
This shared displacement arises from estimating the reference, not from changing the model. Appendix~\ref{app:protocol} gives its expected squared magnitude.

\subsection{External check on CC-1000}\label{sec:cc1000}
The external CC-1000 check uses a frozen 1,000-item Civil Comments subset with a binary toxicity task \citep{borkan2019}. We sample 1,000 training comments from the official Kaggle Jigsaw Unintended Bias archive \citep{jigsaw2019}, with 200 items in each of five toxicity-fraction strata, distinct articles, and at least 50 human annotations per item. The sample contains 176,251 human votes, with a median of 67 annotations per item (range 50--2,196). Each item's empirical reference is $h_i=(1-p_i,p_i)$, where $p_i$ is its human toxicity fraction. All 32 judges have valid labels on all 1,000 items. The panel has the same nominal size as the ChaosNLI panel but different members and family composition, so this is not a controlled comparison of datasets.

Table~\ref{tab:cc1000} summarizes the results. The spectral effective size is $\nu_H=2.838$, while the empirical-reference error-matched size is $\nu_{\mathrm{MSE}}=1.490$. A finite-annotation sensitivity calculation gives 1.549 under the conditional-independence assumptions in the supplementary protocol; it is not the same target as the empirical-reference match. The consensus-direction variance share is $\gamma_{\mathrm{co}}=57.1\%$: over half of the centered residual variance lies along the direction retained by equal-weight averaging. The binary-error diagnostic gives $n_{\mathrm{eff}}=1.677$ using the thresholded human majority, with 10 tied items excluded.

\begin{table}[t]
\centering\small
\setlength{\tabcolsep}{4pt}
\begin{tabular}{@{}lr@{}}
\toprule
Measurement & CC-1000 \\
\midrule
$\nu_H$ & 2.838 \\
$\nu_{\mathrm{MSE}}$ (empirical) & 1.490 \\
$\nu_{\mathrm{MSE}}$ (finite annotation) & 1.549 \\
$n_{\mathrm{eff}}$ & 1.677 \\
$\gamma_{\mathrm{co}}$ & 57.1\% \\
\bottomrule
\end{tabular}
\caption{Measurements for the external CC-1000 check.}
\label{tab:cc1000}
\end{table}

The separation persists on a binary task with a different panel: $\nu_H$ is $1.90\times$ $\nu_{\mathrm{MSE}}$, and the consensus-direction share exceeds one-half.

\subsection{Selecting a panel in the fixed pool}\label{sec:panel_search_eval}
\begin{figure}[t]
\centering
\includegraphics[width=\linewidth]{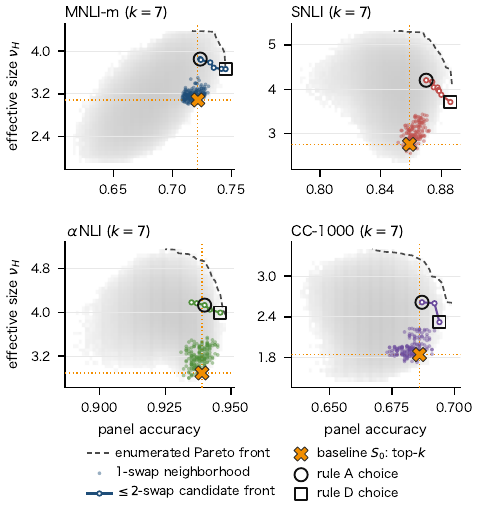}
\caption{Enumeration and candidate panels in the $(\mathrm{acc},\nu_H)$ plane at $k=7$.}\label{fig:search_front_k7}
\end{figure}

\paragraph{Setup.}
We enumerate every panel of size $k=5$ and $k=7$ on the three ChaosNLI tasks and on CC-1000. This gives $\binom{32}{5}=201{,}376$ and $\binom{32}{7}=3{,}365{,}856$ panels per dataset; for each panel we compute $\mathrm{acc}$ and analytically calibrated $\nu_H$. We use odd $k$ because binary majority votes can tie only at even $k$: the panel-level mean tie rate is $0$--$1.3\%$ at $k\in\{5,7\}$ against $2.1\%$--$12.3\%$ at $k\in\{4,8\}$. The baseline $S_0$ is the top-$k$ panel by single-judge accuracy, and the candidate set is the at-most-two-swap neighborhood, $1.81\%$ of all panels at $k=5$ and $0.19\%$ at $k=7$. Effective sizes in this section come from the analytic calibration of Equation~\eqref{eq:nu_closed}, which, unlike the Monte-Carlo grid, is defined below two independent draws; the two CC-1000 baselines fall there, at $\nu_H=1.724$ and $1.843$. Figure~\ref{fig:search_front_k7} shows the full $(\mathrm{acc},\nu_H)$ plane at $k=7$; Figure~\ref{fig:search_front_k5} in the appendix is its $k=5$ counterpart. The distributional readings of Result 3 come from the Gram identity of Equation~\eqref{eq:loss_gram} evaluated on that same candidate set, so the three rules are compared on identical panels with no further sampling.

\paragraph{Result 1: the top-$k$ baseline is dominated in all eight cases.}
Panels strictly more accurate than $S_0$ exist in all eight dataset--size cases (four datasets $\times$ two sizes $k$), between $109$ and $577{,}093$ of them, and $98.55\%$--$100\%$ of those also have a larger $\nu_H$. The joint-improvement set $D(S_0)$ of Section~\ref{sec:panel_search} is thus nonempty in every case, and in this pool an accuracy gain almost never costs effective size. The Pareto fronts are stronger still: their most extreme dominating panel improves accuracy by $0.10$--$0.90$ percentage points and $\nu_H$ by $33.6\%$--$91.4\%$. Figure~\ref{fig:search_front_k7} places the baseline below the Pareto front, with both front and candidate panels in its upper-right joint-improvement region. Its $\nu_H$ ranks at the $0.00$--$47.1$ percentile, and it never lies on the Pareto front. Selecting the most accurate individual models therefore leaves joint gains unrealized in all eight cases.

\paragraph{Result 2: one greedy step improves both targets in all eight cases.}
Table~\ref{tab:panel_search} reports both rules on the at-most-two-swap candidate set. Rule A improves accuracy by $0.10$--$1.10$ percentage points and $\nu_H$ by $24.8\%$--$56.0\%$ in all eight cases; rule D improves accuracy by $0.70$--$2.70$ percentage points and $\nu_H$ by $10.8\%$--$40.6\%$. Although it inspects under $2\%$ of the panels, rule D attains the exact enumeration-maximal accuracy in four of the eight cases and falls $0.10$--$0.50$ points short in the rest, while rule A reaches $77.0\%$--$92.0\%$ of the largest $\nu_H$ available at that size. Two swaps deliver larger gains, while one swap already suffices to dominate the baseline. Between $13$ and $118$ of the one-swap panels ($9.6\%$--$67.4\%$ of $N_1$) lie in $D(S_0)$, and rule A restricted to $N_1$ raises $\nu_H$ by $11.9\%$--$29.5\%$ at $0.10$--$0.85$ percentage points more accuracy in all eight cases, roughly half the two-swap gain. Every one-swap panel is dominated by a two-swap candidate, so the candidate front contains only two-swap panels. Under rule A the models swapped in rank $15$--$32$ by single-judge accuracy while those swapped out rank $1$--$6$ (over both rules, $11$--$32$ in and $1$--$7$ out), and under rule A the mean squared inner product $\bar q$ falls by $35.4\%$--$71.4\%$ while mean member energy $\bar\omega$ rises by $2.8\%$--$32.2\%$. The improvement comes from removing mutually redundant strong models, not from adding stronger ones.

\paragraph{Result 3: the panels chosen for spectral size also recover the distribution better.}
Rule A also improves distribution recovery in all eight cases: $\mathcal E$ falls by $10.1\%$--$24.9\%$ and $\nu_{\mathrm{MSE}}$ rises by $11.3\%$--$33.2\%$ relative to $S_0$ (Table~\ref{tab:panel_search_target}). Rule D improves $\nu_{\mathrm{MSE}}$ in all eight cases as well, by $7.5\%$--$29.6\%$, and beats rule A on this target in one of them. Rule E directly minimizes $\mathcal E$ under the same accuracy constraint, achieving the candidate set's optimal $\nu_{\mathrm{MSE}}$, $15.4\%$--$34.4\%$ above $S_0$. Rule A reaches $96.0\%$--$100\%$ of that optimum, which is $70.6\%$--$100\%$ of the improvement it makes available, and in three of the eight cases the two rules return the same panel. Rule E reaches $92.2\%$--$100\%$ of the largest feasible $\nu_H$ and lies inside $D(S_0)$ of Equation~\eqref{eq:joint_improvement} in every case. Mean member energy moves upward: $\bar\omega$ \emph{rises} by $2.8\%$--$32.2\%$ under rule A, so these panels use members that individually fit $h$ worse. Writing Equation~\eqref{eq:loss_gram} as $\mathcal E=\bar\omega/k+k^{-2}\sum_{a\neq b}K_{a,b}$, the fall in the signed cross terms outweighs that rise in every case (Appendix~\ref{app:selection_protocol}). The two objectives remain distinct, yet spectral selection improves recovery in every case; its smallest share of the available $\nu_{\mathrm{MSE}}$ gain is $70.6\%$, on $\alpha$NLI at $k=5$.

\begin{table}[t]
\centering\small
\setlength{\tabcolsep}{2.4pt}
\begin{tabular}{@{}llrrrrrr@{}}
\toprule
& & \multicolumn{2}{c}{Baseline top-$k$} & \multicolumn{2}{c}{Rule A} & \multicolumn{2}{c}{Rule D} \\
\cmidrule(lr){3-4}\cmidrule(lr){5-6}\cmidrule(lr){7-8}
Dataset & $k$ & acc & $\nu_H$ & $\Delta$acc & $\Delta\nu_H$ & $\Delta$acc & $\Delta\nu_H$ \\
\midrule
MNLI-m & 5 & 0.7325 & 2.732 & $+0.80$ & $+27.6$ & $+1.00$ & $+10.8$ \\
MNLI-m & 7 & 0.7215 & 3.085 & $+0.20$ & $+24.8$ & $+2.35$ & $+18.7$ \\
SNLI & 5 & 0.8685 & 2.495 & $+0.25$ & $+56.0$ & $+1.50$ & $+40.6$ \\
SNLI & 7 & 0.8590 & 2.759 & $+1.10$ & $+52.3$ & $+2.70$ & $+34.5$ \\
$\alpha$NLI & 5 & 0.9380 & 2.651 & $+0.20$ & $+43.9$ & $+0.80$ & $+23.8$ \\
$\alpha$NLI & 7 & 0.9390 & 2.890 & $+0.10$ & $+43.0$ & $+0.70$ & $+38.2$ \\
CC-1000 & 5 & 0.6850 & 1.724 & $+0.30$ & $+51.6$ & $+1.40$ & $+32.6$ \\
CC-1000 & 7 & 0.6860 & 1.843 & $+0.10$ & $+41.9$ & $+0.80$ & $+26.0$ \\
\bottomrule
\end{tabular}
\caption{Accuracy and $\nu_H$ changes for panels selected by rules A and D.}
\label{tab:panel_search}
\end{table}

\begin{table}[t]
\centering\small
\setlength{\tabcolsep}{2.4pt}
\begin{tabular}{@{}llrrrrrr@{}}
\toprule
& & \multicolumn{2}{c}{Baseline top-$k$} & \multicolumn{3}{c}{$\Delta\nu_{\mathrm{MSE}}$ (\%)} & \\
\cmidrule(lr){3-4}\cmidrule(lr){5-7}
Dataset & $k$ & $\mathcal E$ & $\nu_{\mathrm{MSE}}$ & A & D & E & $\Delta\bar\omega_A$ \\
\midrule
MNLI-m & 5 & 0.2135 & 2.125 & $+19.9$ & $+7.5$ & $+19.9$ & $+5.7$ \\
MNLI-m & 7 & 0.2066 & 2.196 & $+13.4$ & $+14.6$ & $+15.4$ & $+4.4$ \\
SNLI & 5 & 0.1367 & 2.487 & $+31.0$ & $+29.6$ & $+31.2$ & $+26.6$ \\
SNLI & 7 & 0.1317 & 2.581 & $+33.2$ & $+22.9$ & $+34.4$ & $+18.9$ \\
$\alpha$NLI & 5 & 0.0600 & 2.768 & $+11.3$ & $+10.6$ & $+15.9$ & $+32.2$ \\
$\alpha$NLI & 7 & 0.0588 & 2.825 & $+17.2$ & $+16.1$ & $+18.6$ & $+22.7$ \\
CC-1000 & 5 & 0.2971 & 1.184 & $+28.3$ & $+19.0$ & $+28.3$ & $+4.1$ \\
CC-1000 & 7 & 0.2890 & 1.217 & $+23.7$ & $+15.6$ & $+23.7$ & $+2.8$ \\
\bottomrule
\end{tabular}
\caption{Distributional recovery for panels selected by rules A, D, and E.}
\label{tab:panel_search_target}
\end{table}

\section{Conclusion}\label{sec:conclusion}
A 32-judge panel has target-specific human-equivalent size. On the three ChaosNLI tasks, distributional-error matching gives $\nu_{\mathrm{MSE}}=2.304$, $3.750$, and $3.445$, while spectral matching gives $\nu_H=4.242$, $6.459$, and $6.499$; the binary-error diagnostic remains near two effective votes. The exact identity explains the gap: recovery error depends on member energy and residual orientation, whereas PR retains only normalized spectral concentration. The equal-energy hard-label construction and the CC-1000 check ($\nu_H=2.84$ versus empirical-reference $\nu_{\mathrm{MSE}}=1.490$) demonstrate this distinction.

With the observed moments held fixed, the distributional-error asymptotes are $2.392$, $3.990$, and $3.655$, and 32 judges already reach 96.3\%, 94.0\%, and 94.2\% of them. In this pool, member choice matters more than panel extension: exhaustive enumeration finds joint accuracy--$\nu_H$ improvements in all eight dataset--size cases, while a search over at most two swaps yields 24.8\%--56.0\% higher $\nu_H$ at 0.10--1.10 percentage points higher accuracy. The selected panels improve distribution recovery in all eight cases and attain 96.0\%--100\% of the candidate set's optimal $\nu_{\mathrm{MSE}}$ under the same accuracy constraint.

\subsection*{Limitations}
Our primary evidence covers three categorical language-inference datasets and one fixed model pool. CC-1000 is a score-only external check with a different panel and a high-annotation, stratified sample. The empirical human distributions have finite-label sampling error, and the fixed-pool asymptote holds the observed squared mean residual, member variance, and covariance fixed when extrapolating beyond 32 judges; a different extension of the same moments shifts it slightly, and neither version bounds what selection or reweighting could reach. The reported quantities measure fixed responses relative to specified targets; they are not general human-replacement rates, out-of-pool selection guarantees, or majority-vote error bounds, and the selected panels are evaluated on the items they were chosen on. Presentation records include cache reuse, and panel additions overlap in judges and items, so those diagnostics are descriptive rather than independent replications.

\subsection*{Data Availability}
The vote archive and the score-only CC-1000 release are available from the public vote repository \citep{li2026votes}.

\FloatBarrier
\bibliography{references}

\FloatBarrier\appendix
\renewcommand{\thefigure}{\thesection\arabic{figure}}
\renewcommand{\thetable}{\thesection\arabic{table}}
\renewcommand{\theequation}{\thesection\arabic{equation}}
\setcounter{figure}{0}\setcounter{table}{0}\setcounter{equation}{0}
\Needspace{310pt}
\section{Collection and Analysis Protocols}\label{app:protocol}

\subsection{Model panel}
Table~\ref{tab:judge_panel} lists the fixed model pool. Requested service identifiers and collection limits are documented in the vote supplement.
\par\smallskip\noindent
\begin{minipage}{\linewidth}
\centering\small
\captionsetup{type=table}
\setlength{\tabcolsep}{3pt}
\begin{tabularx}{\linewidth}{@{}l >{\raggedright\arraybackslash}X@{}}
\toprule
Provider & Models \\
\midrule
OpenAI (5) & GPT5.6-sol, GPT5.6-terra, GPT5.4, o4-mini, GPT4.1 \\
Alibaba (4) & Qwen3.8-max, Qwen3.7-plus, Qwen3.5-plus, Qwen3-max \\
Google (4) & Gemini3.7-flash, Gemini3.6-flash, Gemini3.1-pro, Gemini2.5-pro \\
Moonshot (4) & KimiK3, KimiK2.7-code, KimiK2.6, KimiK2.5 \\
Zhipu (4) & GLM5.3, GLM5.2, GLM5.1, GLM5 \\
Anthropic (3) & Claude-opus5, Claude-sonnet4.6, Claude-haiku4.5 \\
DeepSeek (3) & DeepSeekV4-pro, DeepSeekV4-flash, DeepSeekV3.2 \\
ByteDance (2) & Doubao-seed2.0-pro, Doubao-seed1.8 \\
xAI (2) & Grok4.6, Grok4.5 \\
MiniMax (1) & MiniMaxM3 \\
\bottomrule
\end{tabularx}
\caption{The fixed panel of 32 model judges.}\label{tab:judge_panel}
\end{minipage}
\FloatBarrier

\subsection{Prompts and response sets}
All tasks use temperature 0, one user message, and no system message. MNLI-m and SNLI share the following prompt:
\begin{samepage}
\begin{Prompt}
Given the following premise and hypothesis, determine the relationship between them.

Premise: {premise}
Hypothesis: {hypothesis}

What is the relationship? Reply with ONLY one word: "entailment", "neutral", or "contradiction".
\end{Prompt}
\end{samepage}
The $\alpha$NLI prompt is:
\begin{samepage}
\begin{Prompt}
Given two observations and two hypotheses, determine which hypothesis better explains what happened between the two observations.

Observation 1: {obs1}
Observation 2: {obs2}
Hypothesis 1: {hyp1}
Hypothesis 2: {hyp2}

Which hypothesis is more plausible? Reply with ONLY one word: "1" or "2".
\end{Prompt}
\end{samepage}
Presentation variants reorder the three label names or exchange the two abductive hypotheses. Responses are mapped back to canonical labels. Parsing accepts the legal labels and specified aliases; unparseable responses are flagged. Baseline placeholders are retained as stated in Section~\ref{sec:protocol}; presentation comparisons require parseable paired responses.

The initial 500 presentation items are the first 500 rows of the stored entropy-stratified sample, ordered by original dataset row; the procedure uses no second random item sample. MNLI-m and SNLI require all 31 retained models to be parseable in the original and two reordered prompt versions; $\alpha$NLI requires all 29 models in the original and first reordered version. This leaves 489, 492, and 492 items. Baseline and presentation analyses use separately stored response snapshots. Presentation analyses use the original prompt version for every member of the original panel rather than the full baseline matrix. Figure~\ref{fig:presentation} compares the original and first reordered versions on the common all-variant item set. Model and item lists accompany the protocol and item manifest.

\subsection{Human simulation and anchor size}
Each grid point uses 12 direct Monte Carlo replicates. Inverting each replicate's curve separately gives sample standard deviations of $\nu_H$ of 0.0052, 0.0134, and 0.0203 for the three full panels. These describe simulation spread only. Replicate random numbers are reused across grid sizes and anchor configurations. All nine mean reference curves for three tasks and three anchors are strictly increasing.

The mode anchor in Table~\ref{tab:anchor_comparison} selects the first maximal count in the order entailment, neutral, contradiction (or option 1, option 2). Its tie rule differs from the supplied $\mathrm{gold}_i$ on 6, 2, and 2 items.

\begin{figure}[tbp]
\centering
\includegraphics[width=\linewidth]{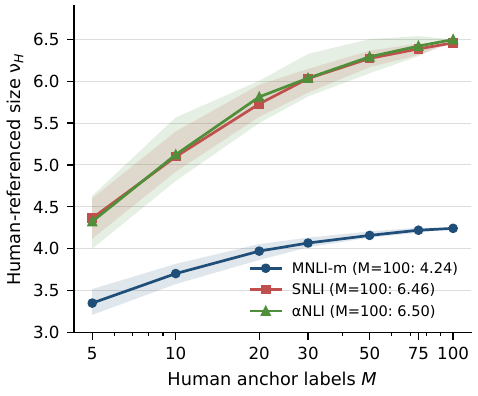}
\caption{Human-reference label count $M$ and the resulting $\nu_H$.}\label{fig:anchor_humans}
\end{figure}

For $M=5,10,20,30,50,75,100$, draw $M$ labels without replacement from each item's 100-label pool and normalize to obtain $A_{M,i}$. Each partial-count setting uses 60 anchor replicates; each replicate recomputes the observed PR and reference curve. Simulated labels always come from full $h_i$, and both model and simulated residuals subtract $A_{M,i}$. Figure~\ref{fig:anchor_humans} shows medians and 5th--95th percentiles; $M=100$ is deterministic.

Conditional on the finite label pool, the shared anchor-estimation displacement has expected squared magnitude
\begin{equation}\label{eq:anchor_mse}
\begin{aligned}
&\mathbb E_{A\mid h}\left[\frac1n\sum_i\|A_{M,i}-h_i\|^2\right]\\
&\qquad=\frac{100-M}{99M}\frac1n\sum_i(1-\|h_i\|^2).
\end{aligned}
\end{equation}
It decreases to zero at $M=100$. This finite-pool statement is not an error estimate relative to an unknown human population.

\subsection{Subpanel sampling and search}
For 9-judge diverse panels, omit one family with probability proportional to $1/s_g$, where $s_g$ is its size, then select one member uniformly from each remaining family. This samples uniformly over valid panels. For size 16, require at least one member from every family and minimize $\sum_g\binom{b_g}{2}$ over family allocations $b_g$. Weight optimal allocations by $\prod_g\binom{s_g}{b_g}$ and sample uniformly without replacement within families. This again samples uniformly over valid panels. Each design uses 2,000 draws.

The search in Figure~\ref{fig:subpanel_search} uses sizes $2,4,6,8,12,16,24,32$. At fixed size, minimizing $Q(I)=\sum_{a,b\in I,\ a\ne b}C_{a,b}^2$ maximizes $\mathrm{PR}_I=|I|^2/(|I|+Q(I))$. Start from the model with smallest squared-inner-product row sum and from 20 random starts. Add members greedily by the smallest increase in $Q$, then apply improving one-in/one-out swaps and keep the best candidate.

Panel search and evaluation use all items in Figure~\ref{fig:subpanel_search}a. For Figure~\ref{fig:subpanel_search}b, 50 random half splits generate candidates on the first half and rebuild the human reference for evaluation on the second half. The full panel is a candidate. Because the second half also selects the size, these gains are exploratory rather than independent test performance.

\subsection{Independent-transition reference}
For each model, estimate a transition matrix $T_a$ from paired responses, conditional on the original label. Independently sample each item's replacement from the corresponding row, leaving other models unchanged. An unobserved original-label row is defined as no change. This matches conditional flip rates and destination frequencies in expectation.

Each model uses 200 replicates. The bands in Figure~\ref{fig:presentation}a are their 2.5th--97.5th percentiles, without multiplicity correction. Figure~\ref{fig:presentation}b plots
$|\mathbb E_{\mathrm{sim}}\Delta\bar q_{\mathrm{sim}}|/
|\Delta\bar q_{\mathrm{swap}}|$,
which is undefined for a zero denominator.

\subsection{Selection protocol and verification materials}\label{app:selection_protocol}
This subsection fixes the rules behind Section~\ref{sec:panel_search_eval}, Tables~\ref{tab:panel_search} and \ref{tab:panel_search_target}, and Figures~\ref{fig:search_front_k7} and \ref{fig:search_front_k5}.

\paragraph{Accuracy and ties.} Panel accuracy never breaks a tie. With $\bar p_i$ the equal-weight label frequencies and $M_i=\{l:\bar p_i[l]\ge\max_l\bar p_i[l]\}$ the set of modal labels, the item score is $\mathbf1[\mathrm{gold}_i\in M_i]/|M_i|$, the expectation under uniform tie-breaking. Because $|M_i|\in\{1,2,3\}$, scores scaled by $\operatorname{lcm}(1,2,3)=6$ are integers, so panel accuracy is an exact integer sum and comparisons between candidates carry no floating-point ties. The task majority vote in Appendix~\ref{app:protocol} resolves ties by a deterministic hash for fixed-panel readings; selection results use the rule above. A tie among all $L$ labels requires $L\mid k$, so odd $k$ eliminates ties only for $L=2$: at $L=3$ two-way ties need no such divisibility and remain possible ($2$--$2$--$1$ at $k=5$, $3$--$3$--$1$ at $k=7$) while three-way ties are identically zero for $k\in\{4,5,7,8\}$. Measured on the baseline panels, tie rates are $1.1\%$, $0.3\%$, $0$, and $0$ at $k=5$ and $0.6\%$, $0.2\%$, $0$, and $0$ at $k=7$, against $12.4\%$, $7.3\%$, $2.1\%$, and $7.0\%$ at $k=4$ and $6.2\%$, $3.2\%$, $2.1\%$, and $4.0\%$ at $k=8$ on MNLI-m, SNLI, $\alpha$NLI, and CC-1000. Gold labels are used as supplied.

\paragraph{Baseline and candidate set.} $S_0$ takes the $k$ highest single-judge accuracies against $\mathrm{gold}_i$; exact ties are ordered by model name, a rule that does not bind at the cut in any of the eight cases. The candidate set is $N_1\cup N_2$ with $|N_1|=k(32-k)$ and $|N_2|=\binom k2\binom{32-k}2$, giving $135+3{,}510$ at $k=5$ and $175+6{,}300$ at $k=7$. Rules A and E share the feasible set $\{\mathrm{acc}>\mathrm{acc}(S_0)\}$, of size $85$, $908$, $282$, $4{,}311$, $773$, $981$, $576$, and $449$ in the eight cases; within it, $85$, $906$, $282$, $4{,}258$, $756$, $981$, $555$, and $439$ candidates also raise $\nu_H$ and lie in $D(S_0)$. Restricted to $N_1$, the panels dominating $S_0$ number $13$, $43$, $14$, $118$, $79$, $60$, $39$, and $25$. Rule A compares $\nu_H$, rule E compares $\mathcal E$, and rule D compares accuracy and then $\nu_H$. A tie after a rule's comparisons goes to the first candidate in enumeration order, which lists $N_1$ before $N_2$, each ordered by the members removed and then by those added, and so prefers the smaller edit. Accuracy is compared as an exact integer sum; $\nu_H$ and $\mathcal E$ are compared as double-precision values with no tolerance. Effective sizes use the analytic calibration of Equation~\eqref{eq:nu_closed}; $\mathcal E$ uses Equation~\eqref{eq:loss_gram}.

\paragraph{Error decomposition.} Writing $\mathcal E=\bar\omega/k+B$ with the signed cross term $B=k^{-2}\sum_{a\neq b}K_{a,b}$ of Equation~\eqref{eq:loss_gram}, rule A increases $\bar\omega/k$ by $0.0016$--$0.0121$ and lowers $B$ by $0.0121$--$0.0687$ relative to $S_0$ in all eight cases, so the fall in $B$ outweighs the rise in member energy. On MNLI-m at $k=5$, the changes are $+0.0044$ and $-0.0398$, giving $\Delta\mathcal E=-0.0354$. Because $\bar q$ averages squared normalized inner products, it discards the signs retained by $B$; a fall in $\bar q$ alone does not imply a fall in $\mathcal E$.

\paragraph{Selected panels.} Table~\ref{tab:selected_members} gives $S_0$ and the swaps for the three rules in all eight dataset--size cases. Parenthesized numbers are single-judge accuracy ranks.

\begin{table}[t]
\centering
\scriptsize
\setlength{\tabcolsep}{3pt}
\begin{tabularx}{\linewidth}{@{}l c >{\raggedright\arraybackslash}X@{}}
\toprule
Dataset & $k$ & Baseline and selected panels (parenthesized values are single-judge accuracy ranks) \\
\midrule
MNLI-m & 5 & \textbf{$S_0$}: KimiK3, KimiK2.5, o4-mini, KimiK2.6, Qwen3.7-plus; \textbf{A}: $-$KimiK2.5(2), $-$Qwen3.7-plus(5), $+$GPT5.6-terra(18), $+$Claude-haiku4.5(24); \textbf{D}: $-$o4-mini(3), $-$Qwen3.7-plus(5), $+$Claude-opus5(13), $+$GPT5.6-terra(18); \textbf{E}: same as A. \\
\addlinespace
MNLI-m & 7 & \textbf{$S_0$}: KimiK3, KimiK2.5, o4-mini, KimiK2.6, Qwen3.7-plus, Qwen3-max, GLM5.3; \textbf{A}: $-$KimiK2.5(2), $-$Qwen3.7-plus(5), $+$DeepSeekV3.2(15), $+$Claude-haiku4.5(24); \textbf{D}: $-$Qwen3-max(6), $-$GLM5.3(7), $+$GPT5.6-terra(18), $+$Claude-haiku4.5(24); \textbf{E}: $-$KimiK2.5(2), $-$Qwen3-max(6), $+$DeepSeekV3.2(15), $+$Claude-haiku4.5(24). \\
\addlinespace
SNLI & 5 & \textbf{$S_0$}: Gemini3.1-pro, Claude-opus5, Qwen3.7-plus, Qwen3.8-max, GLM5.1; \textbf{A}: $-$Gemini3.1-pro(1), $-$Qwen3.8-max(4), $+$Claude-haiku4.5(30), $+$DeepSeekV3.2(32); \textbf{D}: $-$Claude-opus5(2), $-$Qwen3.8-max(4), $+$GPT5.4(21), $+$DeepSeekV3.2(32); \textbf{E}: $-$Qwen3.7-plus(3), $-$Qwen3.8-max(4), $+$KimiK2.6(23), $+$GPT5.6-terra(29). \\
\addlinespace
SNLI & 7 & \textbf{$S_0$}: Gemini3.1-pro, Claude-opus5, Qwen3.7-plus, Qwen3.8-max, GLM5.1, Gemini3.6-flash, Qwen3.5-plus; \textbf{A}: $-$Gemini3.1-pro(1), $-$Qwen3.8-max(4), $+$Claude-haiku4.5(30), $+$DeepSeekV3.2(32); \textbf{D}: $-$Qwen3.8-max(4), $-$Gemini3.6-flash(6), $+$GLM5.2(11), $+$GPT5.6-terra(29); \textbf{E}: $-$Qwen3.8-max(4), $-$Gemini3.6-flash(6), $+$Claude-haiku4.5(30), $+$DeepSeekV3.2(32). \\
\addlinespace
$\alpha$NLI & 5 & \textbf{$S_0$}: Grok4.6, Grok4.5, Qwen3.8-max, Claude-opus5, Gemini3.1-pro; \textbf{A}: $-$Grok4.5(2), $-$Gemini3.1-pro(5), $+$GPT5.6-terra(24), $+$DeepSeekV3.2(32); \textbf{D}: $-$Grok4.5(2), $-$Qwen3.8-max(3), $+$DeepSeekV4-flash(12), $+$MiniMaxM3(21); \textbf{E}: $-$Grok4.5(2), $-$Qwen3.8-max(3), $+$Gemini2.5-pro(18), $+$o4-mini(25). \\
\addlinespace
$\alpha$NLI & 7 & \textbf{$S_0$}: Grok4.6, Grok4.5, Qwen3.8-max, Claude-opus5, Gemini3.1-pro, Gemini3.6-flash, Gemini3.7-flash; \textbf{A}: $-$Grok4.5(2), $-$Gemini3.6-flash(6), $+$GPT5.6-terra(24), $+$DeepSeekV3.2(32); \textbf{D}: $-$Gemini3.6-flash(6), $-$Gemini3.7-flash(7), $+$GPT5.6-terra(24), $+$DeepSeekV3.2(32); \textbf{E}: $-$Gemini3.6-flash(6), $-$Gemini3.7-flash(7), $+$o4-mini(25), $+$DeepSeekV3.2(32). \\
\addlinespace
CC-1000 & 5 & \textbf{$S_0$}: Doubao-seed1.8, GPT5.5, GLM5.2, Claude-opus5, Qwen3.8-max; \textbf{A}: $-$Doubao-seed1.8(1), $-$Claude-opus5(4), $+$Claude-haiku4.5(21), $+$KimiK2.7-code(30); \textbf{D}: $-$Claude-opus5(4), $-$Qwen3.8-max(5), $+$GLM5.1(18), $+$Claude-haiku4.5(21); \textbf{E}: same as A. \\
\addlinespace
CC-1000 & 7 & \textbf{$S_0$}: Doubao-seed1.8, GPT5.5, GLM5.2, Claude-opus5, Qwen3.8-max, Doubao-seed2.0-pro, GPT4.1; \textbf{A}: $-$Claude-opus5(4), $-$Doubao-seed2.0-pro(6), $+$Claude-haiku4.5(21), $+$KimiK2.6(31); \textbf{D}: $-$Doubao-seed2.0-pro(6), $-$GPT4.1(7), $+$Qwen3.7-plus(15), $+$Claude-haiku4.5(21); \textbf{E}: same as A. \\
\bottomrule
\end{tabularx}
\caption{Baseline and selected panels for the eight dataset--size cases.}
\label{tab:selected_members}
\end{table}

\subsection{Binary-error size and majority vote}

Define $\varepsilon_{a,i}=\mathbf1[l(a,i)\ne\mathrm{gold}_i]$ and let $\bar\phi$ be the mean pairwise Pearson correlation of these errors. Table~\ref{tab:effective_sizes} uses
$n_{\mathrm{eff}}=k/[1+(k-1)\bar\phi]$
\citep{kish1965,kohli2026b}, requiring nonzero error variances and positive denominator. This design effect describes averaging standardized errors; it directly gives a variance ratio for raw mean errors only under equal error variances.

Task majority vote selects among tied labels using a deterministic hash of the item and complete response vector. The geometric $J_{\mathrm{maj}}$ instead selects the first modal label in canonical order. Results use their respective definitions. Exact requested model identifiers, seeds, and tie algorithms are recorded in the protocol.

\setcounter{figure}{0}\setcounter{table}{0}\setcounter{equation}{0}
\section{Geometric References and Projections}\label{app:geometry}

\subsection{Individual residual geometry}\label{sec:residual_geometry}
For three-label tasks, the $2\times2$ second moment $G_a=B_{a,a}$ describes an individual judge's residual geometry. With eigenvalues $\sigma_{a,1}^2\geq\sigma_{a,2}^2$ and positive trace, define anisotropy
\begin{equation}\label{eq:anisotropy}
A_a=\frac{\sigma_{a,1}^2-\sigma_{a,2}^2}
{\sigma_{a,1}^2+\sigma_{a,2}^2}\in[0,1].
\end{equation}
Larger $A_a$ means greater concentration along one axis. The principal-axis angle $\theta_a$ is defined modulo $\pi$ relative to $v_1$ \citep{mardia2000}; it is unidentified when the eigenvalues coincide. Binary residuals are one-dimensional, so we do not apply this geometry to $\alpha$NLI.

We compare each judge with simulated label sequences drawn itemwise from $h$. The analytic human second moment sets the angular origin. A Mahalanobis distance measures displacement in angle--anisotropy coordinates relative to the simulated cloud, accounting for their scale and covariance. 

Two constructed judges clarify these displacements. $J_{\mathrm{gold}}$ returns $\mathrm{gold}_i$ on each item without model inference, providing a comparison inspired by negative controls \citep{lipsitch2010}. $J_{\mathrm{maj}}$ returns the panel's modal label, resolving ties in the canonical order of Section~\ref{sec:residual}. It is used for geometry; task-agreement results use the separate majority-vote tie rule in Appendix~\ref{app:protocol}.

\subsection{Human geometry and cloud distance}
Let $V=[v_1,v_2]$. For itemwise human-reference labels $Y_i\sim h_i$, the expected uncentered residual second moment is
\begin{equation}\label{eq:geometry_reference}
G_*=\frac1n\sum_i V^\top[\operatorname{diag}(h_i)-h_ih_i^\top]V.
\end{equation}
If $w_a$ is the leading unit eigenvector of $G_a$, its axial angle is $\theta_a=\operatorname{atan2}(w_{a,2},w_{a,1})\bmod\pi$. Applying the same definition to $G_*$ gives $\theta_*$ and reference anisotropy $A_*$. The plotted difference is
\begin{equation}\label{eq:geometry_angle}
\Delta\theta_a=((\theta_a-\theta_*+\pi/2)\bmod\pi)-\pi/2.
\end{equation}
It is converted to degrees and is invariant to eigenvector sign. Repeated leading eigenvalues leave the axis unidentified.

For a simulated sequence $b$, set $z_b=((180/\pi)\Delta\theta_b,A_b)^\top$. The mean of 6,400 simulated points is $\mu_0$, and their sample covariance, with denominator 6,399, is $\Sigma_0$. Assuming positive definiteness,
\begin{equation}\label{eq:geometry_distance}
d(z)=\sqrt{(z-\mu_0)^\top\Sigma_0^{-1}(z-\mu_0)}.
\end{equation}
Ellipses at $d=1,2$ are equal-distance contours, not confidence regions of specified coverage. The angle origin uses the analytic reference; distances use the empirical cloud center.

With $\mu_{A,0}$ the center's anisotropy, the gold-to-model excess ratio is
$(A_{J_{\mathrm{gold}}}-\mu_{A,0})/
(k^{-1}\sum_a A_a-\mu_{A,0})$,
requiring nonzero denominator. It compares displacement magnitudes and is not an additive fraction of explained error.

\subsection{Projecting the consensus direction}
Figure~\ref{fig:consensus_projection} selects the first model name in lexicographic order within each family: Doubao-seed1.8, DeepSeekV3.2, Gemini2.5-pro, GLM5, GPT4.1, Grok4.5, Claude-haiku4.5, KimiK2.5, MiniMaxM3, and Qwen3.5-plus.

Let $P_2$ contain the first two orthonormal eigenvectors of $S$, and set $w=P_2^\top u_1=(w_1,w_2)^\top$. For $\|w\|>0$, the displayed directions are
\begin{equation}\label{eq:geometry_projection}
u_\parallel=\frac{P_2w}{\|w\|},\qquad
u_\perp=\frac{P_2(-w_2,w_1)^\top}{\|w\|}.
\end{equation}
They form an orthonormal basis of the displayed plane, but $u_\parallel$ need not equal the full $u_1$. The projection is undefined at $w=0$; a tie between the second and third eigenvalues makes the plane nonunique. Plot axes have equal scale and limits at the 99th percentile of $\max(|\mathrm{PC1}|,|\mathrm{PC2}|)$; a small number of points lie outside the frame. The displayed PC scores in Figure~\ref{fig:consensus_projection} use residual coordinates uniformly rescaled by $\sqrt{2}$ before projection; this does not change directions or normalized variance shares.

\setcounter{figure}{0}\setcounter{table}{0}\setcounter{equation}{0}
\section{Spectral Size and Distributional Error}\label{app:loss}

\subsection{Representation and summary}
For a symmetric positive semidefinite unit-diagonal matrix $R$ with $k\geq2$, let $\bar r$ and $\overline{r^2}$ be the mean and mean square over unordered off-diagonal pairs. Signed aggregation $k/[1+(k-1)\bar r]$, when its denominator is positive, and PR $k/[1+(k-1)\overline{r^2}]$ summarize different moments. Table~\ref{tab:crossed_representations} crosses these summaries with the binary-error Pearson matrix $\Phi$ and human-residual Gram matrix $C$. The signed summary of $\Phi$ is $n_{\mathrm{eff}}$. Applying the same expression to $C$ is descriptive and is not a separate human-count equivalence.

Moving from $\Phi$ to $C$ changes the representation, including encoding, reference, and centering: Pearson correlations center binary errors, whereas $C$ retains the means of distributional residuals. Table~\ref{tab:anchor_comparison} isolates the anchor, varying it within the same uncentered residual construction and recalibrating each.

\begin{table}[!htbp]
\centering\small
\setlength{\tabcolsep}{6pt}
\begin{tabular}{@{}lrr@{}}
\toprule
Dataset & Signed summary & PR \\
\midrule
\multicolumn{3}{c}{Binary-error Pearson matrix $\Phi$} \\
\midrule
MNLI-m & 1.971 & 3.536 \\
SNLI & 2.227 & 4.383 \\
$\alpha$NLI & 1.999 & 3.792 \\
\midrule
\multicolumn{3}{c}{Human-residual Gram matrix $C$} \\
\midrule
MNLI-m & 2.201 & 4.228 \\
SNLI & 2.877 & 6.422 \\
$\alpha$NLI & 2.736 & 6.427 \\
\bottomrule
\end{tabular}
\caption{Dependence summaries under binary-error and human-residual representations.}\label{tab:crossed_representations}
\end{table}

\subsection{Error identities}
For a one-hot label $Y\sim h_i$, $\mathbb EY=h_i$ and $\operatorname{tr}\operatorname{Cov}(Y)=1-\|h_i\|^2$. Averaging $m$ conditionally independent labels divides this variance by $m$. Averaging over items yields $\mathcal E_H(m)=J/m$ and hence Equation~\eqref{eq:loss_human}. This expectation concerns within-item draws and does not require independence across items.

Decompose the raw Gram matrix as
$K_{a,b}=\sum_t(S_t)_{a,b}+\mu_a^\top\mu_b$.
Substitution into $k^{-2}\mathbf1^\top K\mathbf1$ yields Equation~\eqref{eq:loss_consensus}. The covariances use denominator $n$ and the original residual scale; member-normalized covariances cannot replace them.

\paragraph{Two ways to hold moments fixed.}\label{app:ceiling_moments}
The fixed-pool asymptote of Equation~\eqref{eq:pool_ceiling} depends on which moments the extension preserves. Write $\bar\omega=k^{-1}\operatorname{tr}K$ and $\bar c_K$ for the mean off-diagonal entry of $K$, so that $\bar c_K=(k\mathcal E-\bar\omega)/(k-1)$ at $k=32$. For a uniformly drawn subset $I$ of $m$ members from the fixed pool,
\begin{equation}\label{eq:subset_error_exact}
\mathbb E_{I:|I|=m}[\mathcal E(I)]=\bar c_K+\frac{\bar\omega-\bar c_K}{m},
\end{equation}
which is exact for $1\leq m\leq k$: it follows from averaging $m^{-2}\mathbf1^\top K_I\mathbf1$ over subsets, and we verify it against exhaustive enumeration at $m=2$ and $m=3$. Extending that curve holds the uncentered moments fixed and gives the limit $\bar c_K$, whereas Equation~\eqref{eq:pool_ceiling} extends the centered curve and holds $\mu_2$, $\bar v$, and $\bar\rho$ fixed. With $\delta_\mu=k^{-1}\sum_a\|\mu_a-\bar\mu\|^2$,
\begin{equation}\label{eq:ceiling_gap}
\bar c_K=\bar c_S+\mu_2-\frac{\delta_\mu}{k-1},
\qquad
\mathcal E_\infty=\bar c_K+\frac{\delta_\mu}{k-1},
\end{equation}
where $\bar c_S$ is the mean off-diagonal centered covariance summed over channels. The two limits therefore differ by $\delta_\mu/(k-1)$: $3.97\times10^{-4}$, $2.74\times10^{-4}$, and $1.74\times10^{-5}$ in $\mathcal E$ on MNLI-m, SNLI, and $\alpha$NLI, or $0.21\%$, $0.32\%$, and $0.04\%$, which moves $\nu_{\mathrm{MSE},\infty}$ from $2.392$, $3.990$, and $3.655$ to $2.397$, $4.003$, and $3.656$. The observed panels reach over $93\%$ of either limit. Both limits describe extensions of the observed second moments; neither bounds what selection, reweighting, or a different class of judges could achieve, and $\delta_\mu$ is computable from each member's mean label frequencies alone, since subtracting the same reference cancels in differences between members.

\subsection{Spectral orientation identity}
Under the positive-energy assumption, $K=DCD$ and $D\mathbf1=d$. Substituting an orthonormal eigendecomposition gives
\begin{equation}\label{eq:orientation_proof}
\begin{aligned}
\mathcal E
&=\frac{d^\top Cd}{k^2}
=\frac1{k^2}\sum_j\lambda_j(w_j^\top d)^2\\
&=\frac{\Omega}{k^2}\sum_j\lambda_jb_j.
\end{aligned}
\end{equation}
Parseval's identity gives $\sum_jb_j=1$. For a repeated eigenvalue, individual $b_j$ depend on the basis within its eigenspace, but their sum over that eigenspace and the weighted sum in Equation~\eqref{eq:loss_orientation} do not. The identity decomposes the observed error; PR alone cannot predict it. It uses the uncentered Gram matrix; replacing it by centered covariance would drop the mean-residual term.

\subsection{Realizable hard-label counterexamples}
Equal spectra can produce different errors. With four items and $h_i=(1/2,1/2)$, encode labels by signs and set $a=(1,1,-1,-1)^\top$, $b=(1,-1,1,-1)^\top$. Panels $[a,a,b,b]$ and $[a,-a,b,-b]$ both have normalized Gram eigenvalues $(2,2,0,0)$, PR of 2, and member energies $1/2$. Their distributional errors are $1/4$ and 0. What differs is the eigenvector orientation relative to equal-weight averaging.

The nonnegative-correlation construction in Section~\ref{sec:recovery} uses rows as items and columns as judges:
\begin{equation}\label{eq:patterns}
\begin{aligned}
Z&=\begin{pmatrix}1&1&1&1\\1&1&-1&-1\\-1&-1&1&1\\-1&-1&-1&-1\end{pmatrix},\\
H&=\begin{pmatrix}1&1&1&1\\1&-1&1&-1\\1&1&-1&-1\\1&-1&-1&1\end{pmatrix}.
\end{aligned}
\end{equation}
Form $Z_A=[Z;Z]$ and $Z_B=[H;\mathbf1_4\mathbf1_4^\top]$, then append each matrix's negation: $[Z_A;-Z_A]$ and $[Z_B;-Z_B]$. Both panels now have 16 items with the same $h_i=(1/2,1/2)$. Mapping signs to one-hot labels gives energy $1/2$ for every member and zero across-item mean residual. Their normalized Gram matrices are
\begin{equation}\label{eq:nonnegative}
\begin{aligned}
C^{(A)}&=\begin{pmatrix}1&1&0&0\\1&1&0&0\\0&0&1&1\\0&0&1&1\end{pmatrix},\\
C^{(B)}&=\tfrac12I_4+\tfrac12\mathbf1_4\mathbf1_4^\top.
\end{aligned}
\end{equation}
Panel $A$ has $(\bar c_C,v_C,\bar q)=(1/3,2/9,1/3)$; panel $B$ has $(1/2,0,1/4)$. Equations~\eqref{eq:pr} and \eqref{eq:loss_equal_energy} give PR $2\to16/7$ and error $1/4\to5/16$. The conflict requires neither negative correlations, unequal energies, nor an across-item mean displacement.

\setcounter{figure}{0}\setcounter{table}{0}\setcounter{equation}{0}
\section{Presentation and Ranking Diagnostics}\label{app:diagnostics}

\subsection{Individual agreement and error clustering}
Agreement with $\mathrm{gold}_i$ ranges from 59.7\% to 72.0\% on MNLI-m, 71.2\% to 85.3\% on SNLI, and 84.4\% to 93.5\% on $\alpha$NLI; the corresponding model means are 66.0\%, 81.2\%, and 90.6\%. Rankings differ by task: Grok4.6 has the lowest agreement on MNLI-m and the highest on $\alpha$NLI. Panel majority vote achieves 68.6\%, 86.5\%, and 93.0\%, exceeding the best individual model only on SNLI.

\begin{figure}[!htbp]
\centering
\includegraphics[width=0.88\linewidth]{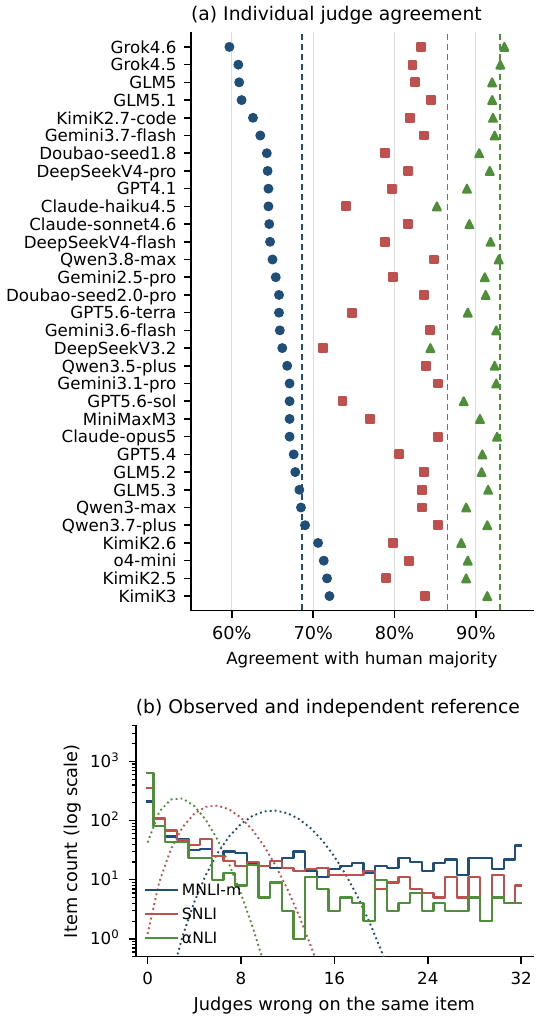}
\caption{(a) Individual agreement with human gold. (b) Observed error counts and an unconditional-independence reference.}\label{fig:judges}
\end{figure}

Errors also cluster across judges (Figure~\ref{fig:judges}). All 32 judges disagree with gold on 38, 8, and 4 items, respectively. A reference that multiplies the judges' marginal error probabilities predicts far fewer than one such item. This comparison rejects the unconditional independence description of these data; item-difficulty heterogeneity can also produce clustering, so it does not identify a specific source of shared model error.

\subsection{Exploratory search in the fixed pool}\label{sec:exploratory_search}
\begin{figure}[tbp]
\centering
\includegraphics[width=0.88\linewidth]{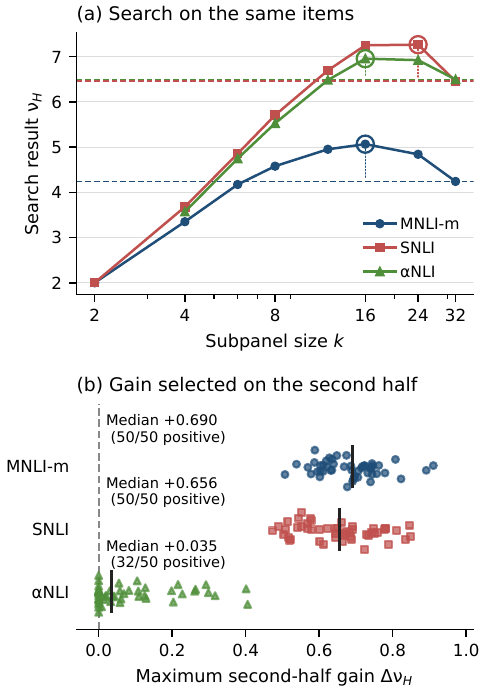}
\caption{Subpanel search in the fixed model pool. (a) Best candidates by size. (b) Half-split search results.}\label{fig:subpanel_search}
\end{figure}

Larger panels do not always have larger $\nu_H$. Among the searched candidates in Figure~\ref{fig:subpanel_search}a, MNLI-m and $\alpha$NLI peak at $k=16$, reaching 5.067 and 6.958. SNLI peaks at $k=24$ with 7.268, close to 7.254 at $k=16$. All full 32-model panels lie below these peaks. This is a property of the searched candidates in this pool, distinct from the trend in random-panel medians.

Figure~\ref{fig:subpanel_search}b uses 50 half-item splits. Candidates are generated at each size on one half, then evaluated on the other half, where the final size is also selected. Median gains over the full panel are 0.690, 0.656, and 0.035. Because the full panel is a candidate and the second half also selects the size, these gains are exploratory search diagnostics rather than held-out selection benefits.

\subsection{Provider families}\label{sec:families}
To test family membership as a diversity proxy, we compare 2,000 random 9-judge panels with 2,000 panels containing at most one judge per family. Family diversity yields small median $\nu_H$ gains on SNLI and $\alpha$NLI and almost no change on MNLI-m (Table~\ref{tab:families}). Its value depends on the task even within the same model pool.

\begin{table}[t]
\centering\small
\setlength{\tabcolsep}{5pt}
\begin{tabular}{@{}lrrr@{}}
\toprule
Dataset & Random & Diverse & Difference \\
\midrule
MNLI-m & 3.384 & 3.382 & -0.003 \\
SNLI & 4.509 & 4.615 & +0.105 \\
$\alpha$NLI & 4.468 & 4.577 & +0.108 \\
\bottomrule
\end{tabular}
\caption{Median $\nu_H$ for random and family-diverse 9-judge panels.}\label{tab:families}
\end{table}

\subsection{Additional presentation diagnostics}\label{app:presentation_diagnostics}
Section~\ref{sec:presentation} and Figure~\ref{fig:presentation} compare observed presentation changes with independent transitions. We also examine changes in spectral size and the destinations of changed labels within the same saved responses.

For 200 random 10-judge subpanels, replacing each member once gives 2,000 comparisons per dataset. Mean changes in $\nu_H$ are $+0.006$, $-0.014$, and $+0.096$. The observed replacements provide no consistent cross-task increase in $\nu_H$.

\paragraph{Flip destinations.}
The fraction of originally incorrect responses among flips is 1.62, 2.81, and 5.36 times the overall error rate among valid model--item pairs. Pooling 1,524 MNLI-m flips and 1,127 SNLI flips, 87.4\% and 91.7\% go to the highest-human-support label after excluding the original label; ties take the first label in canonical order. For binary $\alpha$NLI, removing the original leaves no destination choice.

The originally correct subsets have 701 and 549 flips, with corresponding destination rates 90.3\% and 92.5\%. The originally incorrect subsets have 823 and 578 flips, with rates 84.9\% and 90.8\%, equal to their rates of switching to gold. These data therefore do not distinguish choosing a well-supported alternative from choosing gold among originally incorrect responses.

On $\alpha$NLI, flips favor the first-presented option for 22 judges and the other option for 7. The median judge-level fraction selecting the first-presented option after a flip is 61.5\%, whereas pooling all flips gives 53.9\%.

\begin{figure}[t]
\centering
\includegraphics[width=0.88\linewidth]{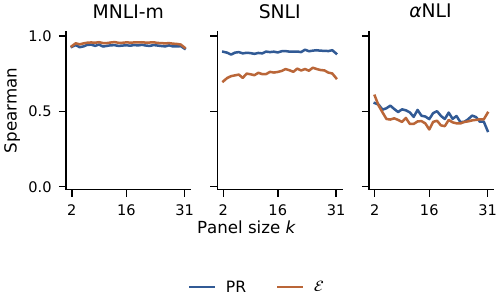}
\caption{Within-size rank correlations between fixed item halves for PR and distributional error $\mathcal E$.}\label{fig:rank_stability}
\end{figure}

\subsection{Within-size comparisons and item stability}
For each $k=2,\ldots,31$, we draw 2,000 panels without replacement within a panel but with possible repeats across draws. Deduplicating member sets and including the full 32-judge panel leaves 54,139 panels per dataset. There is no within-size correlation at $k=32$, where only one panel exists. The protocol supplies seeds and member lists.

For $k=4,8,16,24,31$, adding each absent model gives 150,372 original-panel/added-member records per dataset. Different additions can yield the same enlarged panel and share both judges and items. Item-identifier hashes define two fixed 500-item groups; raw Gram matrices and PR are recomputed within each group. The split lists accompany the supplement.

Table~\ref{tab:loss_additions} requires the same conflict direction on the full set and both halves, with both absolute relative changes at least 1\% on every set. Numerical ties are excluded at $|\Delta\mathrm{PR}|\leq10^{-10}$ or $|\Delta\mathcal E|\leq10^{-12}$. These overlapping records describe the fixed pool and are not independent replications.

\paragraph{Failed-item sensitivity.}\label{app:failure_sensitivity}
Removing the 1/0/5 affected items on MNLI-m, SNLI, and $\alpha$NLI changes full-panel $\mathcal E$ from 0.196911 to 0.197088 on MNLI-m and from 0.048254 to 0.048357 on $\alpha$NLI; SNLI is unchanged. For $4\to5$, stable conflicts passing the same 1\% filter change from 2,116/309/1,127 to 2,079/309/1,055 out of 54,348 additions per task. The filtered views retain the original item-half memberships without refilling removed items. These saved sensitivity results preserve the presence of conflicts and their task-dependent frequencies. The selection results of Section~\ref{sec:panel_search_eval} use the placeholder-retained items. Rerunning the whole selection, including $\delta$ and the full enumeration, on the filtered views leaves $S_0$, all 18 ChaosNLI A/D/E panels, and every enumeration count unchanged; the largest shifts are $0.004$ percentage points in accuracy gain and $0.34$ points in $\nu_H$ gain, and the largest $S_0$ percentile moves from $47.1$ to $47.4$. CC-1000 has no placeholder labels.

Figure~\ref{fig:rank_stability} reports rank agreement between item halves across all panel sizes. On $\alpha$NLI, both PR and error rankings are less stable than on MNLI-m, demonstrating sensitivity to item composition even within a fixed model pool.

\subsection{Full item populations}\label{app:population}
The main analyses use 1,000 items per ChaosNLI task, drawn as 334, 333, and 333 items from the three human-entropy terciles of the full task, cut on ChaosNLI's entropy field. On MNLI-m and SNLI each tercile holds one-third of the population, so this allocation is proportional; on $\alpha$NLI the population shares are 28.5\%, 36.3\%, and 35.2\%.

A later collection round scored every item of the three tasks with a 32-judge panel that shares 26 models with the fixed panel and replaces the other six. Items with any unparseable response are dropped, leaving 1,596, 1,513, and 1,525 of 1,599, 1,514, and 1,532 items. Where the two rounds share models and items, all 77,844 votes coincide. We run the unchanged pipeline, including code, reference grid, 12 replicates, seed, and selection rules, on this panel twice, on the sample items it retains and on its full item sets, so that only the item set differs between the two runs (Table~\ref{tab:population}).

\begin{table}[!htbp]
\centering\small
\setlength{\tabcolsep}{4pt}
\begin{tabular}{@{}lrrrrr@{}}
\toprule
Dataset & Items & $\nu_H$ & $\nu_{\mathrm{MSE}}$ & Ratio & Share \\
\midrule
MNLI-m & 999 & 4.220 & 2.269 & 1.86 & 96.4\% \\
 & 1,596 & 4.242 & 2.278 & 1.86 & 96.4\% \\
SNLI & 1,000 & 6.650 & 3.739 & 1.78 & 94.0\% \\
 & 1,513 & 6.737 & 3.804 & 1.77 & 93.9\% \\
$\alpha$NLI & 995 & 6.283 & 3.335 & 1.88 & 94.5\% \\
 & 1,525 & 6.287 & 3.149 & 2.00 & 94.3\% \\
\bottomrule
\end{tabular}
\caption{Effective sizes for the replacement panel on retained samples and full item sets.}\label{tab:population}
\end{table}

Moving to the full item sets changes $\nu_H$ by $+0.5\%$, $+1.3\%$, and $+0.1\%$ and $\nu_{\mathrm{MSE}}$ by $+0.4\%$, $+1.7\%$, and $-5.6\%$. The spectral size stays 1.86, 1.77, and 2.00 times the error-matched size, and 32 judges attain 96.4\%, 93.9\%, and 94.3\% of the fixed-pool asymptote. The largest change is on $\alpha$NLI, whose allocation is not proportional. Exchanging the six models on the same sample items moves $\nu_H$ by $-0.5\%$, $+3.0\%$, and $-3.3\%$ and $\nu_{\mathrm{MSE}}$ by $-1.6\%$, $-0.3\%$, and $-3.2\%$, the same order as the change of item set.

On the full item sets, enumeration at $k\in\{5,7\}$ finds panels that beat the accuracy-top-$k$ baseline in both accuracy and $\nu_H$ in all six cases (294 to 39,682 panels per case), and the baseline is never on the Pareto front. Swapping at most two members raises $\nu_H$ by 23.5--47.8\% at 0.03--0.26 percentage points higher accuracy. On the retained sample items one case differs: the replacement panel's top-5 on MNLI-m already has the highest accuracy of all 201,376 five-judge panels, so no panel beats it on both, although equal-accuracy panels with higher $\nu_H$ keep it off the Pareto front. All main-text results use the fixed panel on the 1,000-item samples, the configuration of the released votes and the presentation experiments.

\subsection{Candidate fronts at \texorpdfstring{$k=5$}{k=5}}\label{app:search_front_k5}
\begin{figure}[!htbp]
\centering
\includegraphics[width=\linewidth]{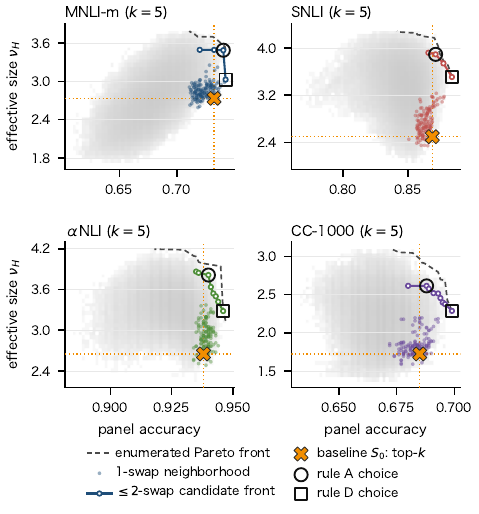}
\caption{Enumeration and candidate panels in the $(\mathrm{acc},\nu_H)$ plane at $k=5$.}\label{fig:search_front_k5}
\end{figure}

Figure~\ref{fig:search_front_k5} repeats Figure~\ref{fig:search_front_k7} of Section~\ref{sec:panel_search_eval} at $k=5$, with identical elements and colors. The candidate set holds $|N_1|=135$ and $|N_2|=3{,}510$ panels, $3{,}645$ in total, which is $1.81\%$ of the $\binom{32}{5}=201{,}376$ enumerated panels. On all four datasets the one-swap neighborhood again hugs the baseline: $13$--$79$ of its $135$ panels dominate $S_0$, yet none of them lies on the at-most-two-swap candidate front, each being dominated by a two-swap candidate. The four $k=5$ rows of Table~\ref{tab:panel_search} give the corresponding values.

\end{document}